%% file: neurips_2024.tex
\documentclass{article}
\PassOptionsToPackage{numbers, compress}{natbib}
\usepackage[preprint]{neurips_2024}

\usepackage[utf8]{inputenc} 
\usepackage[T1]{fontenc}    
\usepackage[pagebackref=true,breaklinks=true,letterpaper=true,colorlinks,urlcolor=black,bookmarks=false]{hyperref}
\hypersetup{citecolor=[RGB]{119,185,0}}

\usepackage{url}            
\usepackage{booktabs}       
\usepackage{amsfonts}       
\usepackage{nicefrac}       
\usepackage{microtype}      
\usepackage{xcolor}         

\usepackage{graphicx}
\usepackage{amsmath,amssymb}
\usepackage{amsthm}
\usepackage{booktabs}
\usepackage{algorithm}
\usepackage{algorithmic}
\usepackage{lineno}
\usepackage{multirow} 
\usepackage{rotating}
\usepackage{enumitem}
\usepackage{color}
\usepackage{marvosym}
\usepackage{booktabs}
\usepackage{bm}
\usepackage{rotating}
\usepackage{enumitem}

\usepackage{colortbl}
\usepackage{array}
\usepackage{makecell}
\usepackage{svg}
\usepackage{wrapfig}
\usepackage{subcaption}
\usepackage{adjustbox}

\usepackage{arydshln}
\usepackage{multicol}

\definecolor{tablecolor}{gray}{.93}
\definecolor{demphcolor}{RGB}{144, 144, 144}
\definecolor{boost_line}{RGB}{197,90,17}
\definecolor{adapt_line}{RGB}{75,140,174} 

\definecolor{pink}{RGB}{221, 160, 221} 
\definecolor{red}{RGB}{255, 0, 0} 

\newcommand{\demph}[1]{\textcolor{demphcolor}{#1}}
\newcommand{\descend}[1]{\textcolor{adapt_line}{\textbf{#1}}}
\newcommand{\improve}[1]{\textcolor{boost_line}{\textbf{#1}}}
\newcommand{\rotbox}[1]{\rotatebox{70}{#1}}

\newtheorem{theorem}{Theorem}

 \title{\makebox[\textwidth][c]{Revisiting Prompt Pretraining of Vision-Language Models}}

\author{Zhenyuan Chen$^1$,
Lingfeng Yang$^3$,
Shuo Chen$^4$,
Zhaowei Chen$^5$,
Jiajun Liang$^5$,
Xiang Li$^{2,1}$\thanks{Corresponding author. Team site: https://github.com/IMPlus-PCALab}
\\ [0.15cm]
\footnotesize
$^1$IMPlus, VCIP, College of Computer Science, Nankai University~~
$^2$NKIARl, Shenzhen Futian~~\\
\footnotesize
$^3$Nanjing University of Science and Technology~~
$^4$RIKEN~~
$^5$MEGVII Technology
\\ [0.15cm]
\footnotesize
\{zhenyuanchen424, chaowechan\}@gmail.com,~
yanglfnjust@njust.edu.cn,\\
\footnotesize
shuo.chen.ya@riken.jp,~
liangjiajun@megvii.com,~
xiang.li.implus@nankai.edu.cn}

\begin{document}

\maketitle

\begin{abstract}
    Prompt learning is an effective method to customize Vision-Language Models (VLMs) for various downstream tasks, involving tuning very few parameters of input prompt tokens. Recently, prompt pretraining in large-scale dataset (e.g., ImageNet-21K) has played a crucial role in prompt learning for universal visual discrimination. However, we revisit and observe that the \emph{limited} learnable prompts could face underfitting risks given the \emph{extensive} images during prompt pretraining, simultaneously leading to poor generalization. To address the above issues, in this paper, we propose a general framework termed \textbf{R}evisiting \textbf{P}rompt \textbf{P}retraining (\textbf{RPP}), which targets at improving the fitting and generalization ability from two aspects: prompt structure and prompt supervision. 
    For prompt structure, we break the restriction in common practice where query, key, and value vectors are derived from the shared learnable prompt token. Instead, we introduce unshared individual query, key, and value learnable prompts, thereby enhancing the model's fitting capacity through increased parameter diversity.
    For prompt supervision, we additionally utilize soft labels derived from zero-shot probability predictions provided by a pretrained Contrastive Language Image Pretraining (CLIP) teacher model. These soft labels yield more nuanced and general insights into the inter-class relationships, thereby endowing the pretraining process with better generalization ability.
    RPP produces a more resilient prompt initialization, enhancing its robust transferability across diverse visual recognition tasks. Experiments across various benchmarks consistently confirm the state-of-the-art (SOTA) performance of our pretrained prompts. Codes and models will be made available soon.
\end{abstract}

\section{Introduction}
Due to the remarkable capabilities of large Vision-Language Models (VLMs)~\cite{radford2021learning,li2021align,fang2023eva}, they are widely adopted for visual classification~\cite{zhou2022learning,zhou2022conditional,khattak2023self}, object detection~\cite{gu2021open,li2022grounded}, and semantic segmentation~\cite{zhou2022extract,liang2023open}, etc.
In contrast to the traditional backbone fine-tuning or linear probing of vision-only models~\cite{he2022masked,xie2022simmim}, popular tuning techniques for VLMs include the cross-attention block~\cite{alayrac2022flamingo,li2023blip}, adapter or projector~\cite{gao2023clip,liu2023visual}, and prompt learning~\cite{zhou2022learning,khattak2023self}.
Prompt learning methods have gained prominence within the landscape of VLM tuning techniques due to their convenience and lightweight characteristics. These methods focus extensively on specialized tuning by training distinct prompts tailored for each domain or task~\cite{zhou2022learning,zhou2022conditional,khattak2023maple,khattak2023self,chen2022prompt}. Unfortunately, these specialized prompts, while optimized for specific narrow domains, often lack wider generalization capabilities.

To broaden applicability across diverse downstream domains using prompts, one promising solution is to introduce the \emph{pretraining paradigm} of prompts over extensive images. Recently, \underline{P}r\underline{OM}pt \underline{P}retraining (POMP)~\cite{ren2023prompt} makes the first attempt by pretraining a shared token prompt that captures a wide spectrum of visual concepts on the ImageNet-21K~\cite{deng2009imagenet} dataset. 
This pretraining on a large-scale dataset is crucial for infusing the token prompt with semantic information, thereby facilitating universal visual discrimination. Nevertheless, we observe that the limited set of learnable prompts (i.e., very few learnable parameters) may exhibit susceptibility to underfitting when confronted with the copious image data encountered during prompt pretraining, in both training epoch and data amount (see Sec.~\ref{sec.RPP} for a detailed description). In this study, we introduce a novel framework called Revisiting Prompt Pretraining (RPP) as a solution to tackle the underfitting challenge. RPP is devised to enhance the fitting and generalization capabilities of models by introducing a unique approach focused on two key aspects: prompt structure and prompt supervision.

\begin{wrapfigure}{r}{0.49\textwidth}
    \vspace{-13.5pt}
    \centering
    \captionsetup{font={scriptsize}}
    \includegraphics[width=0.475\textwidth]{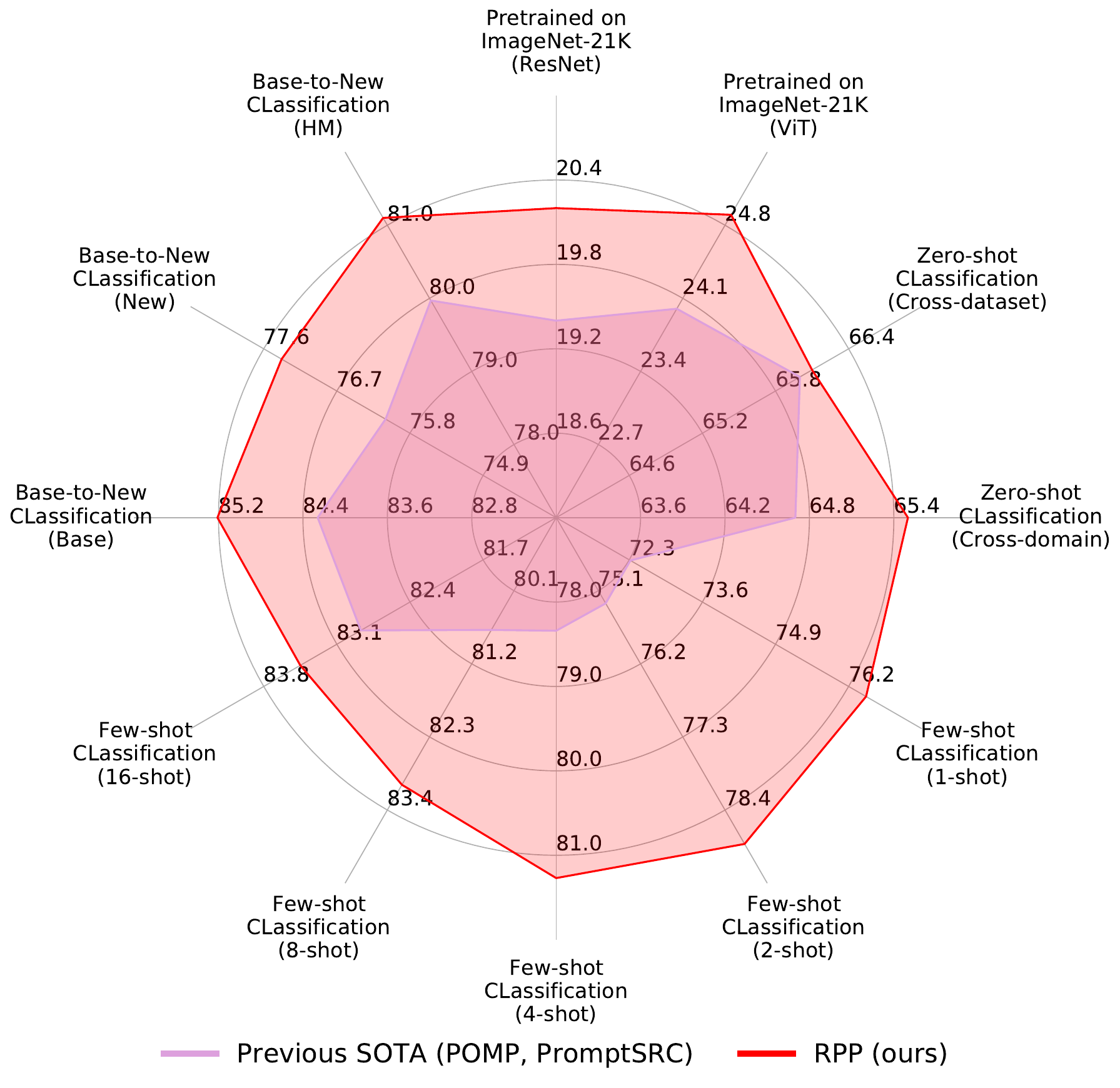}
    \vspace{-6pt}
    \caption{Our method outperforms previous SOTA models on a broad range of visual recognition tasks and datasets.}
    \label{fig_main_result}
    \vspace{-18pt}
\end{wrapfigure}

Regarding prompt structure, we challenge the conventional practice of deriving query, key, and value vectors from a single shared learnable prompt token.
Instead, we adopt a novel approach by introducing unshared individual query, key, and value learnable prompts, thereby increasing the parameter space available for optimization. 
This leads to enhanced fitting capabilities of the model. Particularly, we introduce Self-Attention Prompt Learning (SAPL), a meticulously designed learnable prompt featuring layer-by-layer replaceable query-key-value components. 
However, there is a substantial distribution discrepancy between the prompt pretraining dataset and CLIP training datasets. Fully aligning with the pretraining dataset might compromise the original generalization capability inherent in CLIP, posing a challenge to maintain its broader generalization applicability.

Further, to address the issue of limited generalization ability while simultaneously maintaining a strong pretraining fitting capability, we introduce a regularization technique termed Prompt Pretraining with Knowledge Distillation (PPKD) for prompt supervision. This approach allows for flexible adjustments to visual and textual prompt tokens while preserving robust generalization capabilities from the larger-scale CLIP teacher. Benefiting from the excellent zero-shot generalization ability of the larger-scale CLIP teacher model on the pretraining dataset, we can enhance generalization while simultaneously preserving fitting capabilities for downstream tasks.
Additionally, drawing from previous works such as ~\cite{chen2023boundary,chen2019curvilinear}, we provide theoretical analyses for the generalization ability of RPP. We demonstrate that as the weight assigned to the regularization loss increases, the regularization loss gradually decreases. This convergence allows the model to achieve a reduced upper bound on generalization error, consequently enhancing the overall generalization capability of RPP.

The results presented in Fig.~\ref{fig_main_result} highlight the superior performance of RPP over previous SOTA models across various visual recognition tasks and datasets. Specifically, RPP records a 0.9$\%$ improvement in ImageNet-21K validation accuracy over POMP~\cite{ren2023prompt}.
For zero-shot generalization, RPP shows a 0.43$\%$ enhancement with an average accuracy across 14 datasets.
Notably, in a 16-shot setting, RPP shows enhancements in few-shot and base-to-new generalization accuracies across 11 classification datasets, with improvements of 0.58$\%$ and 1.12$\%$ respectively, compared to PromptSRC~\cite{khattak2023self}.

Our contributions can be summarized as follows: 
\vspace{-5pt}
\begin{itemize}[]
\item[$\bullet$] We are the first to explicitly identify the underfitting issue encountered during Prompt Pretraining in VLMs.
\item[$\bullet$] We propose the SAPL prompt structure and the PPKD prompt supervision to alleviate the underfitting risk while maintaining its generalization. 
\item[$\bullet$] We present a theoretical analysis of generalization ability of RPP, demonstrating a reduction in the upper bound on generalization error that improves overall generalization performance.
\item[$\bullet$] Compared to existing published works, our method consistently achieves SOTA results on multiple datasets under few-shot/base-to-new transfer, with an average improvement of 0.58$\%$ and 1.13$\%$ points. 
\end{itemize}

\begin{figure*}[!h]
    \centering
    \includegraphics[width=1\linewidth]{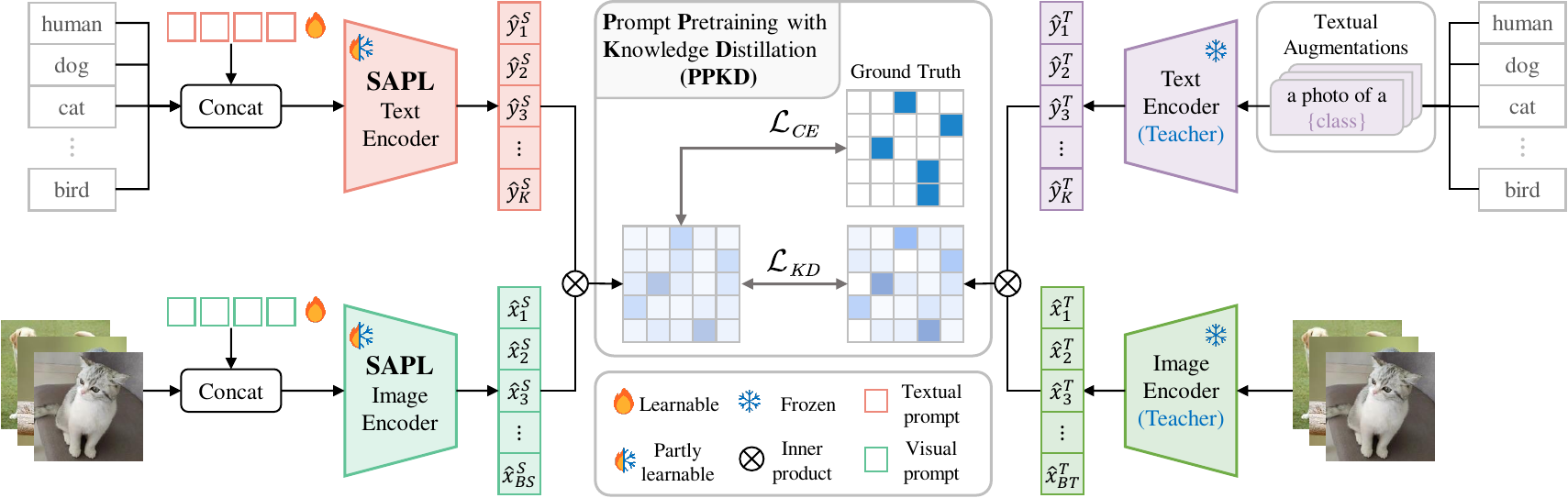}
    \captionsetup{font={footnotesize}}
    \caption{\textbf{An overview of our proposed pretraining framework.} Firstly, we propose SAPL Text/Image Encoder, optimizing individual query, key, and value embeddings directly and explicitly (Sec.~\ref{sec4.2}). Next, we employ a frozen teacher model to supervise the student model's learnable prompts using regularization loss from knowledge distillation (Sec.~\ref{sec4.3}). Further, we provide theoretical results for the generalization error bound of RPP (Sec.~\ref{sec4.4}).}
    \label{fig_main_structure}
    \vspace{-1em}
\end{figure*}

\section{Related Work}
\subsection{Prompt Learning}
The technique of prompt learning is initially introduced as an improvement to manual prompting in the field of Natural Language Processing (NLP) for transferring knowledge from pretrained models to specific downstream domains~\cite{shin2020autoprompt,jiang2020can,zhong2021factual,lester2021power,li2021prefix}.

In recent years, the application of prompt learning has expanded into the realms of vision~\cite{wang2022learning,jia2022visual,bahng2022exploring,huang2023diversity} and vision-language~\cite{zhou2022conditional,khattak2023maple,khattak2023self,rasheed2023fine}.
These approaches involve two main branches: the text branch and the image branch. 
In the text branch, CoOp~\cite{zhou2022learning} employs learnable tokens instead of manually designed suffixes in text inputs, such as ``a photo of a {class}'', to enhance the transfer capabilities in classification tasks.
Subsequently, CoCoOp~\cite{zhou2022conditional} introduces instance-conditional prompts to mitigate model overfitting. These methods adapt to various tasks, including open vocabulary~\cite{feng2022promptdet} and visual grounding~\cite{rao2022denseclip}, etc.
In the image branch, vision prompt tuning serves as an efficient training technique alongside backbone fine-tuning or linear probing. VPT~\cite{wang2022learning} appends learnable tokens before the image sequence but keeps the entire backbone fixed to transfer vision-only models. A concurrent approach~\cite{bahng2022exploring} introduces pixel-wise prompts for spatially structured images instead of sequences. Furthermore, some methods leverage joint textual and visual prompt tuning based on CLIP~\cite{radford2021learning} for improvements~\cite{zang2022unified,khattak2023maple,khattak2023self}.

Moreover, prompt pretraining with large-scale data has been incorporated into enhanced pretrained models, such as POMP~\cite{ren2023prompt}. 
However, upon revisiting, it is observed that the \emph{limited} learnable prompts may face underfitting risks, especially given the \emph{extensive} images during prompt pretraining. 
To address the underfitting issue of POMP, this paper introduces a general framework termed RPP, aiming to enhance fitting and generalization abilities by focusing on two key aspects: prompt structure and prompt supervision, as shown in Fig.~\ref{fig_main_structure}.

\subsection{CLIP Distillation}
Knowledge Distillation (KD)~\cite{hinton2015distilling,heo2019comprehensive} emerges as a common method for transferring knowledge from teacher models with more parameters and better performance to more deployable student models.

Distillation for VLMs has gained prominence in recent years, with efforts to convert VLMs knowledge into vision-only models during model pretraining~\cite{wei2022mvp,fang2023eva,wang2022clip}, open vocabulary detection~\cite{gu2021open,zhong2022regionclip}, or semantic segmentation~\cite{zhou2022extract,jiao2023learning}.
Additionally, some methods directly distill VLMs to VLMs through ordinary distillation~\cite{yang2023clip,fang2021compressing,wu2023tinyclip,li2024promptkd}, self-distillation~\cite{wu2023clipself,andonian2022robust}, and linear probing~\cite{laroudie2023improving}. DistillVLM~\cite{fang2021compressing} transfers knowledge through the intermediate representations of each proposal generated from pretrained detectors. CLIP-KD~\cite{yang2023clip} explores various distillation strategies, including relation, feature, gradient, and contrastive paradigms to assess their impact on CLIP distillation. TinyCLIP~\cite{wu2023tinyclip} learns cross-modal feature alignment from teacher to student in a visual-textual affinity space. CLIPSelf~\cite{wu2023clipself} distills CLIP itself using dense feature maps and corresponding predictions. LP-CLIP~\cite{laroudie2023improving} employs a single linear probing layer for distillation.  

In addition to the hard one-hot labels that may present optimization challenges for specific prompt tokens, we introduce soft labels derived from a pretrained large-scale CLIP teacher's zero-shot probability predictions. 

\section{Method}
\label{sec4}
\subsection{Revisiting Prompt Pretraining}
\label{sec.RPP}
\begin{figure}[!h]
    \centering
    \captionsetup{font={footnotesize}}
    \begin{minipage}[b]{0.45\linewidth}
        \centering
        \includegraphics[width=\linewidth]{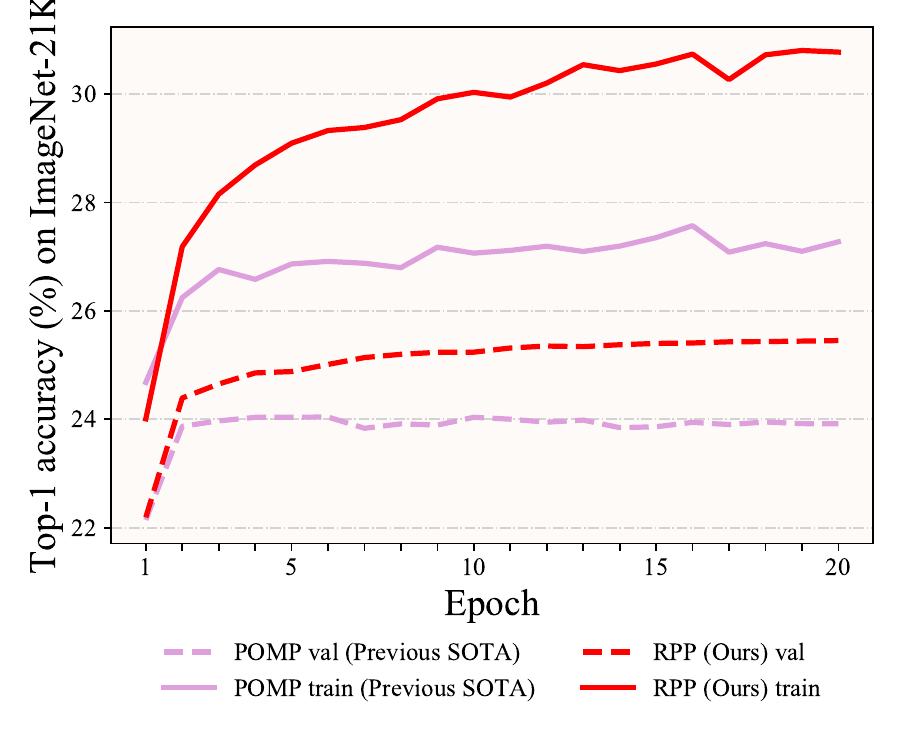}
        \vspace{-2em}
        \subcaption{Comparative pretraining experiments on training epoch. }
        \label{fig_train_val_acc}
    \end{minipage}
    \hspace{0.5em} 
    \begin{minipage}[b]{0.45\linewidth}
        \centering
        \includegraphics[width=\linewidth]{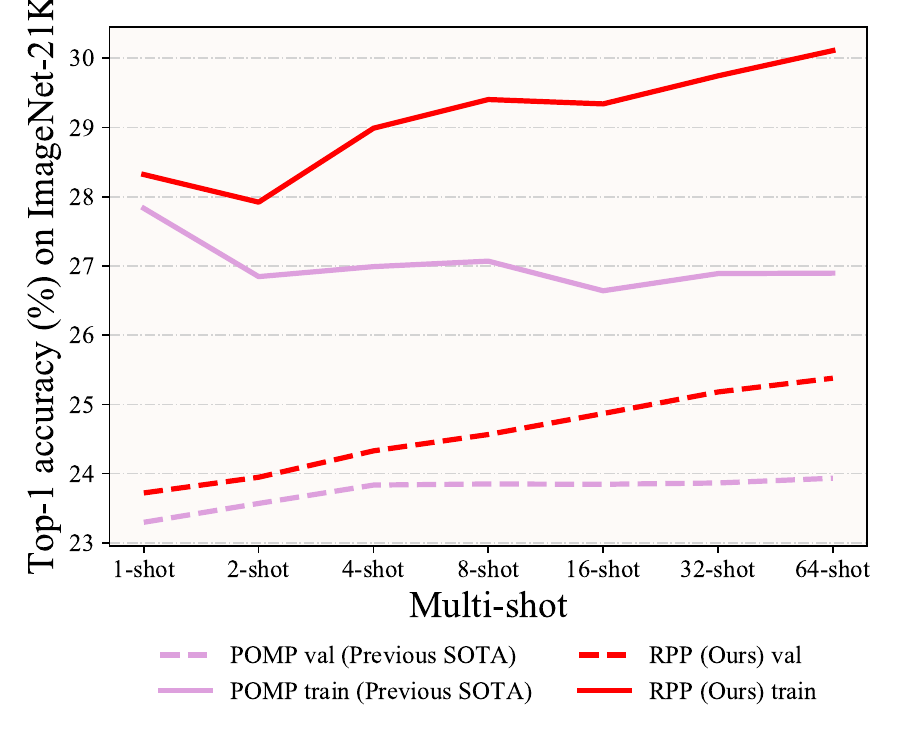}
        \vspace{-2em}
        \subcaption{Comparative pretraining experiments on training data amount.}
        \label{fig_train_val_acc_multishot}
    \end{minipage}
    \caption{\textbf{Empirical evidence of underfitting of current practice.} All methods use the same optimization strategy with 5 epochs and the same data.}
    \label{fig_rpp}
    \vspace{-1em}
\end{figure}

Fig.~\ref{fig_rpp} illustrates the training and testing accuracy of our method compared with the POMP method on the ImageNet-21K dataset, in terms of training epoch and data amount, respectively. The \textcolor{pink}{pink lines} denote the performance of the POMP method, while the \textcolor{red}{red lines} represent our approach. In Fig.~\ref{fig_train_val_acc}, the training and testing accuracy of the POMP method become relatively flat and saturated after around 4 epochs, indicating negligible further fitting to the dataset.
In Fig.~\ref{fig_train_val_acc_multishot}, the POMP method reaches its fitting capacity ceiling at 4-shot, and as training data is further increased, the flattening curve highlights an underfitting issue. In contrast, our method maintains its pretraining fitting capability across extended training epochs and larger data amounts.
A detailed description of the method will be provided in Sec.~\ref{sec4.2} and~\ref{sec4.3}.

\subsection{Preliminaries}
\label{sec4.1}
CLIP~\cite{radford2021learning} contains an image encoder and a text encoder, denoted as $\textbf{ImgEnc}$ and $\textbf{TextEnc}$, respectively. For zero-shot transfer, the input image $I=\left \{ {I_{i}} | {I_{i}} \in \mathbb{R}^{3 \times H \times W},i=1,2,\cdots ,N \right \}$ is embedded into a token sequence and passed through the image encoder, denoted as $\textbf{PatchEmb}$. The corresponding text $T_{c}$ is the categories prepended with a suffix, appearing as ``a photo of a [class]''. $T_{c}$ is tokenized and embedded into a feature space via the text encoder, denoted as $\textbf{Tokenizer}$:
\begin{align}
    x_{i} &= \textbf{ImgEnc}(\textbf{PatchEmb}(I_{i})), \\
    y_{c} &= \textbf{TextEnc}(\textbf{Tokenizer}(T_{c})),
\vspace{0pt}
\label{eqn_clip_image_enc}
\end{align}
where $\mathbf{y} =\left \{ \mathbf{y_{c}}|c=1,2,\cdots ,C \right \} $ with $\mathbf{y_{i}} \in \mathbb{R}^{d}$, and $\chi =\left \{ \mathbf{x_{i}}|i=1,2,\cdots ,N \right \} $ with $\mathbf{x_{i}} \in \mathbb{R}^{d}$ represent the global features of the image and text outputs, respectively. Then the normalized embeddings are derived as $\mathbf{\hat{x}_{i} = x_{i} / \left \| x_{i} \right \|_{2} }$ and $\mathbf{\hat{y}_{c} = y_{c} / \left \| y_{c} \right \|_{2} }$. The probabilities concerning each class are calculated as:
\begin{equation}
    s_i = \frac{\mathrm{exp}(\mathbf{\hat{x}_{i}} \cdot \mathbf{\hat{y}_{c}} / \tau)}{ {\textstyle \sum_{j}^{C}} \mathrm{exp}(\mathbf{\hat{x}_{i}} \cdot \mathbf{\hat{y}_j} / \tau)},
    \vspace{0pt}
    \label{eqn_clip_text_enc}
\end{equation}
where $C$ denotes the number of classes, $\left ( \cdot \right ) $ represents the dot product, and $\tau > 0$ is the temperature.

Different from manually designed instructions, prompt learning provides a solution to transfer knowledge through learnable tokens in the form of continuous parameters. There are two aspects to prompt learning: the textual branch and the visual branch. In the first one, following common practices~\cite{zhou2022learning,khattak2023maple},  ``a photo of a [class]'' is replaced with ``$P^t$ [class]'', where $P^t = \{p^t\}_m$, ($m \in \mathbb{N}_{M}$), is the learnable prompt, and $M$ denotes the number of contents.
In terms of visual prompting~\cite{jia2022visual,zang2022unified,khattak2023maple}, another set of learnable tokens is extended after the image patches: $\left \{ I, P^v \right \}$. Then they are processed in place of the original image patches and text inputs.
Subsequently, $\left \{ P^t, P^v \right \} $ are learned to adapt to specific data domains.

In addition to the integration of learnable prompts in the input space, there are methods~\cite{khattak2023maple,jia2022visual} that propose to incorporate tokens in the deep layers of text / vision encoders, known as multi-layer prompt learning. Denoting the output of the $l$-th transformer block in the text encoder as $\left \{ P^t\left ( l \right ), F^t\left ( l \right ) \right \} $, where $P^t\left ( l \right )$ and $F^t\left ( l \right )$ are the intermediate prompt and text features, respectively. Then $P^t\left ( l \right )$ is replaced with learnable prompts, which are additionally initialized in the $l$-th layer of the encoder to introduce learnable tokens in deeper transformer layers. Likewise, within the forward process in the vision encoder, $\left \{ P^v\left ( l \right ), F^v\left ( l \right ) \right \} $ undergoes the same operation.

\subsection{Self-Attention Prompt Learning}
\label{sec4.2}
\begin{figure}[]
	\centering
	\includegraphics[width=1\linewidth]{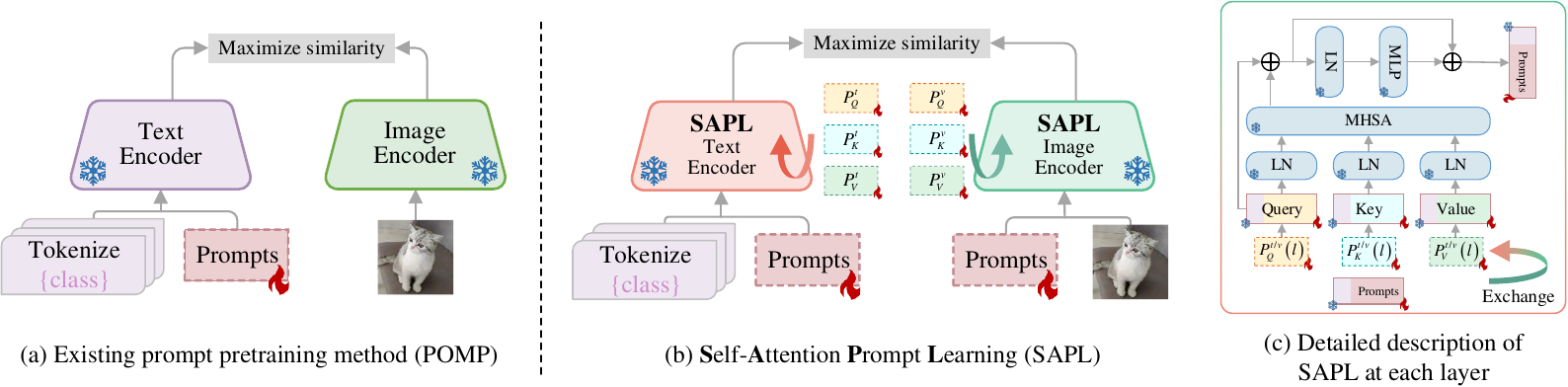}
    \captionsetup{font={footnotesize}}
	\caption{\textbf{A detailed description of our proposed SAPL prompt structure.} (a) Existing methods adopt uni-modal prompting techniques to fine-tune CLIP representations as prompts are learned only in a single branch of CLIP (language or vision). (b) Our SAPL explicitly and directly optimizes the individual query, key, and value embeddings. (c) Detailed description of SAPL at each layer.}
	\label{fig_SaPL_structure}
    \vspace{-1em}
\end{figure}

As shown in Fig.~\ref{fig_SaPL_structure}, to mitigate potential underfitting concerns related to the model architecture, we employ a methodology utilizing independent learnable prompts. These prompts are designated for substitution before the self-attention computation at each layer of the model. Particularly, 
before computing the forward propagation of the current layer with input $X^{t/v}\left ( l \right )=\left \{ P^{t/v}\left ( l \right ),F^{t/v}\left ( l \right ) \right \} $, we utilize $\mathbf{Rep}\left ( \cdot,\cdot \right )$ to replace the learnable prompt from the previous layer with separated learnable prompts $P_{*}^{t/v}\left ( l \right )=\left \{P^{t/v}_{Q}\left ( l \right ), P^{t/v}_{K}\left ( l \right ), P^{t/v}_{V}\left ( l \right ) \right \}$. This approach diverges from traditional methods that typically utilize a single shared prompt, represented by $P^{t/v}\left ( l \right )$. 

It is essential to acknowledge that this method applies when the backbone is ViT~\cite{dosovitskiy2020image}, functioning for both $X^{t}\left ( l \right )$ and $X^{v}\left ( l \right )$. Conversely, for the ResNet~\cite{he2016deep} backbone, it exclusively applies to $X^{t}\left ( l \right )$. This operation aims to broaden the optimization scope within the parameter space while upholding consistent feature scaling. The computation of the current layer is outlined as follows (the following formulas do not distinguish $t/v$):
\begin{equation}
    \left\{\begin{matrix}
    X_{Q}^{*}\left ( l \right )=\mathbf{LN}\left ( \mathbf{Rep}\left (  X\left ( l-1 \right ),P_{Q}\left ( l \right )\right )\right ) ,
     \\
    X_{K}^{*}\left ( l \right )=\mathbf{LN}\left ( \mathbf{Rep}\left (  X\left ( l-1 \right ),P_{K}\left ( l \right )\right )\right ) ,
     \\
    X_{V}^{*}\left ( l \right )=\mathbf{LN}\left ( \mathbf{Rep}\left (  X\left ( l-1 \right ),P_{V}\left ( l \right )\right )\right ) ,
    \end{matrix}\right.
    \vspace{0pt}
    \label{eqn_LN}
\end{equation}
where $l \in \{2, 3, \ldots, 12\}$ represents the $l$-th layer, and $\mathbf{LN}\left ( \cdot\right)$ denotes LayerNorm. We normalize each prompt to maximize the parameter space while simultaneously ensuring that features do not diverge significantly from the original CLIP distribution. 
Then we proceed with the remaining calculations:
\begin{equation}
    \begin{aligned}
    \hat{X}\left ( l \right )&=X_{Q}\left ( l \right )+\mathbf{MHSA}\left (X_{Q}^{*}\left ( l \right )  ,X_{K}^{*}\left ( l \right ) ,X_{V}^{*}\left ( l \right ) \right ),\\
    X\left ( l \right )&=\hat{X}\left ( l \right )+\mathbf{MLP}\left ( \mathbf{LN}\left ( \hat{X}\left ( l \right ) \right ) \right ),
    \end{aligned}
    \vspace{0pt}
    \label{eqn_MHSA}
\end{equation}
where $\mathbf{MHSA}\left (\cdot, \cdot, \cdot \right)$ stands for Multi-head Self Attention,  $\mathbf{MLP}\left (\cdot\right)$ stands for Multi-Layer Perceptron, and $X_{Q}\left ( l \right )$ means exchanged query prompt without layernorm.

\subsection{Prompt Pretraining with Knowledge Distillation}
\label{sec4.3}
As shown in Fig.~\ref{fig_main_structure}, we conduct pretraining of multi-layer prompts by leveraging the extensive ImageNet-21K dataset, pioneering the transfer of robust embedded knowledge from a larger-scale CLIP model to a more computationally efficient scale. As detailed in Sec.~\ref{sec4.2}, we utilize the layer-by-layer replaceable $\left \{ P_{*}^t\left ( l \right ), P_{*}^v\left ( l \right ) \right \} $ as the learnable parameters for the student network, while employing the original pretrained large-scale CLIP model as the teacher network. 

We aim to acquire prompt pretrained CLIP endowed with universal generalization capabilities through disciplined pretraining via distillation on the ImageNet-21K dataset. Nevertheless, direct pretraining on 21K data demands over 300GB of GPU memory~\cite{ren2023prompt}. To address this challenge, we adopt methodologies inspired by POMP~\cite{ren2023prompt}, specifically employing local contrast and local correction techniques. In particular, when presented with a batch of input images, our process involves sampling $K$ classes$\left ( K< < C \right )$. This includes selecting the respective ground-truth class and incorporating $K-1$ randomly selected samples as shared negative classes. Therefore, the probabilities for sample classes are calculated as:
\begin{equation}
    \hat{s}_i = \frac{\mathrm{exp}(\hat{x}_{i} \cdot \hat{y}_{i}^{ram} / \tau)}{\mathrm{exp}(\hat{x}_{i} \cdot y_{i}^{gt} / \tau)+  {\textstyle \sum_{j=1}^{K-1}} \mathrm{exp}(\hat{x} \cdot \hat{y}_{j}^{ram} / \tau + m)},
    \vspace{0pt}
    \label{eqn_local_sim}
\end{equation}
where $y_{i}^{gt}$ denotes the ground-truth class of sample $i$ and $\hat{y}_{i}^{ram}$ denotes the randomly selected $K-1$ negative classes. To address the inherent bias in the prompt optimization direction due to the absence of other negative classes, we introduce a local correction term denoted as $m$ into the probability of negative classes. This term aims to incentivize the positive logit to surpass the negative logits by a predefined margin. The formulation for $m$ is as follows:
\begin{equation}
    m=-\mathrm{log}\left ( \left ( K-1 \right )/\left ( C-1 \right ) \right ).
    \vspace{0pt}
    \label{eqn_local_sim_m}
\end{equation}

Drawing from the aforementioned strategies, we incorporate knowledge distillation loss and cross-entropy loss during the pretraining phase. To enhance text embedding diversity in the teacher model, we employ textual augmentations. This involves the random selection of 60 prompt templates for the text encoder from the comprehensive template list provided in~\cite{radford2021learning}. Then, we introduce logit-level consistency regularization by conditioning the distribution of prompted logits on the teacher logits distribution. This is achieved through the minimization Kullback-Leibler Divergence of the following loss function:
\begin{equation}
    \mathcal{L}_{KD}=\frac{1}{N\cdot K} {\textstyle \sum_{i=1}^{N}}{\textstyle \sum_{k=1}^{K}} s_{ik}^{T}  \mathrm{log}\frac{ s_{ik}^{T}}{s_{ik}^{S}  },
    \vspace{0pt}
    \label{eqn_kd}
\end{equation}
where $\hat{s}_{ik}^{S} \in \left [ 0,1 \right ] $ and $\hat{s}_{ik}^{T} \in \left [ 0,1 \right ] $ represent student's and teacher's probability prediction of $i$-th training data in the $k$-th sampled class. For image classification on the ImageNet-21K dataset $\mathcal{D}$, learnable prompts $\left \{ P_{*}^t\left ( l \right ),P_{*}^v\left ( l \right )  \right \} $ interact with frozen $\textbf{ImgEnc}$ and $\textbf{TextEnc}$ are optimized with the cross-entropy loss, $\mathcal{L}_{CE}$, as:
\begin{equation}
    \mathcal{L}_{CE}=\boldsymbol{\mathbb{E}}_{\left (\hat{x},y^{gt}\right )\sim \mathcal{D}}\mathcal{L}\left ( \hat{s}^{S}\left ( \Theta \right ) ,y^{gt} \right ).
    \vspace{0pt}
    \label{eqn_ce}
\end{equation}

We use $\lambda > 0$ as the loss balancing hyper-parameters. Our overall pretraining objective thus becomes:
\begin{equation}
    \mathcal{L}=\mathcal{L}_{CE}+\lambda\mathcal{L}_{KD}.
    \vspace{0pt}
    \label{eqn_all}
\end{equation}

\subsection{Theoretical Analysis}
\label{sec4.4}
In this section, we provide theoretical results for the generalization error bound of RPP. All proofs of theorems are given in the \textit{appendix} Sec.~\ref{sec.A}.

We define the following optimization objectives according to Eq.~\eqref{eqn_all}:
\begin{equation}
    \min_{\Theta \in \mathbb{R}^{d} } \underset{\mathcal{L}_{CE}}{\underbrace{\frac{1}{N}{\textstyle \sum_{i=1}^{N}}\mathcal{L}\left ( \hat{s}_{i}^{S}\left ( \Theta \right )  ,y_{i}^{gt} \right )} }  +\lambda \underset{\mathcal{L}_{KD}}{\underbrace{\mathcal{L}\left ( \hat{s}^{S}\left ( \Theta \right )  ,\hat{s}^{T} \right ) } } ,
    \vspace{0pt}
    \label{eqn_optim}
\end{equation}
where $\Theta$ represents the set of learning prompts $\left \{ P_{*}^{t},P_{*}^{v} \right \}$ of the proposed network, with $d$ denoting the dimensionality of the learning parameters. Now we further analyze the effectiveness of RPP by offering the generalization error bound. Such a bound evaluates the bias between the generalization error $\varepsilon\left ( \Theta  \right ):=\boldsymbol{\mathbb{E}}_{\left (\hat{s}^{S},y^{gt}\right )\sim \mathcal{D}}\mathcal{L}\left ( \hat{s}^{S}\left ( \Theta  \right ),y^{gt} \right )$ and empirical error $\bar{\varepsilon} _{\chi }\left ( \Theta  \right ):=\frac{1}{N}{\textstyle \sum_{i=1}^{N}}\mathcal{L}\left ( \hat{s}_{i}^{S}\left ( \Theta \right ) ,y_{i}^{gt} \right )$, where $D$ is the real data distribution and $\boldsymbol{\mathbb{E}}\left ( \cdot \right ) $ denotes the expectation function.

\begin{theorem}
    Assume that $\Theta ^{*}$ is the solution to Eq.~\eqref{eqn_optim}. Then we have that for any $0<\delta <1$ with probability $1-\delta $,
    \begin{equation}
        \varepsilon\left ( \Theta^{*}  \right )-\bar{\varepsilon} _{\chi }\left ( \Theta^{*}   \right )\le X^{*} \sqrt{2ln\left ( 1/\delta  \right )/N }+B_{\lambda }~\footnote{Here $ B_{\lambda }=2\mathbb{E}_{\chi,\mathcal{Z}}\left ( \underset{\Theta\in \mathcal{F}\left ( \lambda \right )}{\mathrm{sup}}  \bar{\varepsilon}_{\mathcal{Z}} \left( \Theta \right)-\bar{\varepsilon}_{\chi} \left( \Theta \right) \right )/\mathbb{E}_{\chi,\mathcal{Z}}\left ( \underset{\Theta\in \mathbb{R}}{\mathrm{sup}}  \bar{\varepsilon}_{\mathcal{Z}} \left( \Theta \right)-\bar{\varepsilon}_{\chi} \left( \Theta \right) \right )$ and $\mathcal{F}\left ( \lambda \right )$ is a shrinking hypothesis space induced by the regularizer $\mathcal{L}_{KD}\left ( \Theta  \right )$.}R_{N}\left ( \mathcal{L} \right )~\footnote{Here $R_{N}\left ( \mathcal{L} \right )$ is the Rademacher complexity of the loss function $\mathcal{L}$ related to the space $\mathbb{R}$ for $N$ training examples.}, 
        \vspace{0pt}
        \label{eqn_ged}
    \end{equation}
    where $X^{*}=\mathrm{max}_{r\in \mathbb{N}_{N}}\left |  \mathcal{L}\left( \hat{s}_{r}^{S}\left( \Theta \right), y_{r}^{gt} \right) \right | $, and $B_{\lambda}\to 0$ as $\lambda\to +\infty$. 

\label{theorem.1}
\end{theorem}

In Eq.~\eqref{eqn_ged} the first term of the upper bound converges with the increasing of the number of training data $N$. We can also find that the second term converges to 0 with the increasing of $\lambda$, which means the regularizer $\mathcal{L}_{KD}$ effectively improves the generalization ability of RPP.

\section{Experiments}
The experimental section here presents the extensive results of image classification tasks. For details on ablation studies and experiment configurations, please refer to the \textit{appendix}. 
It is important to highlight that POMP utilizes the ImageNet-21K winter 21 version, requiring the replication of all comparative pretraining experiments on our dataset (ImageNet-21K fall 11 version)\footnote{The complete winter dataset is not available due to server download. All of our implementation experiments will be marked with ``(our impl.)"}.

\subsection{Quantitative Experiments}

\paragraph{\textbf{Prompt Pretraining on ImageNet-21K.}}
\input{tables/pretrain}

The results presented in Tab.~\ref{pretrain-table} showcase the outcomes of pretraining experiments conducted on the ImageNet-21K dataset, demonstrating the superior performance of our proposed method compared to ZeroshotCLIP. Notably, when utilizing ResNet50 and ViT-B/16 as backbones, our method exhibits performance leads of 0.8$\%$ and 0.9$\%$ compared to POMP, respectively. These advancements result from our focused efforts to mitigate underfitting during prompt pretraining, achieved through a refined prompt structure and optimized supervision. 

\input{tables/zero-shot}
\paragraph{\textbf{Zero-shot Image Classification.}}
Tab.~\ref{zero-shot} presents the zero-shot experimental outcomes of our method across ten cross-dataset datasets and four cross-domain datasets. Specifically, Our RPP approach outperforms the POMP method by an average margin of 0.3$\%$ and 0.8$\%$ in cross-dataset and cross-domain tasks, respectively. Compared with the original CLIP framework, our approach yields improvements of 2.4$\%$ and 3.6$\%$. 

It is worth noting that when the downstream dataset distribution closely resembles that of ImageNet-21K, our method benefits from the increased fitting capability introduced by our SAPL prompt structure, resulting in improved generalization performance (e.g., in ImageNet-R, our method shows a 1.1$\%$ performance boost compared to POMP). Conversely, when the downstream dataset distribution significantly differs from ImageNet-21K, our method further enhances the original CLIP's generalization capability, thanks to our PPKD prompt supervision (e.g., in SUN397, our method exhibits a 1.4$\%$ performance improvement compared to POMP).

\input{tables/few-shot}
\begin{figure}[]
    \centering
    \includegraphics[width=1\linewidth]{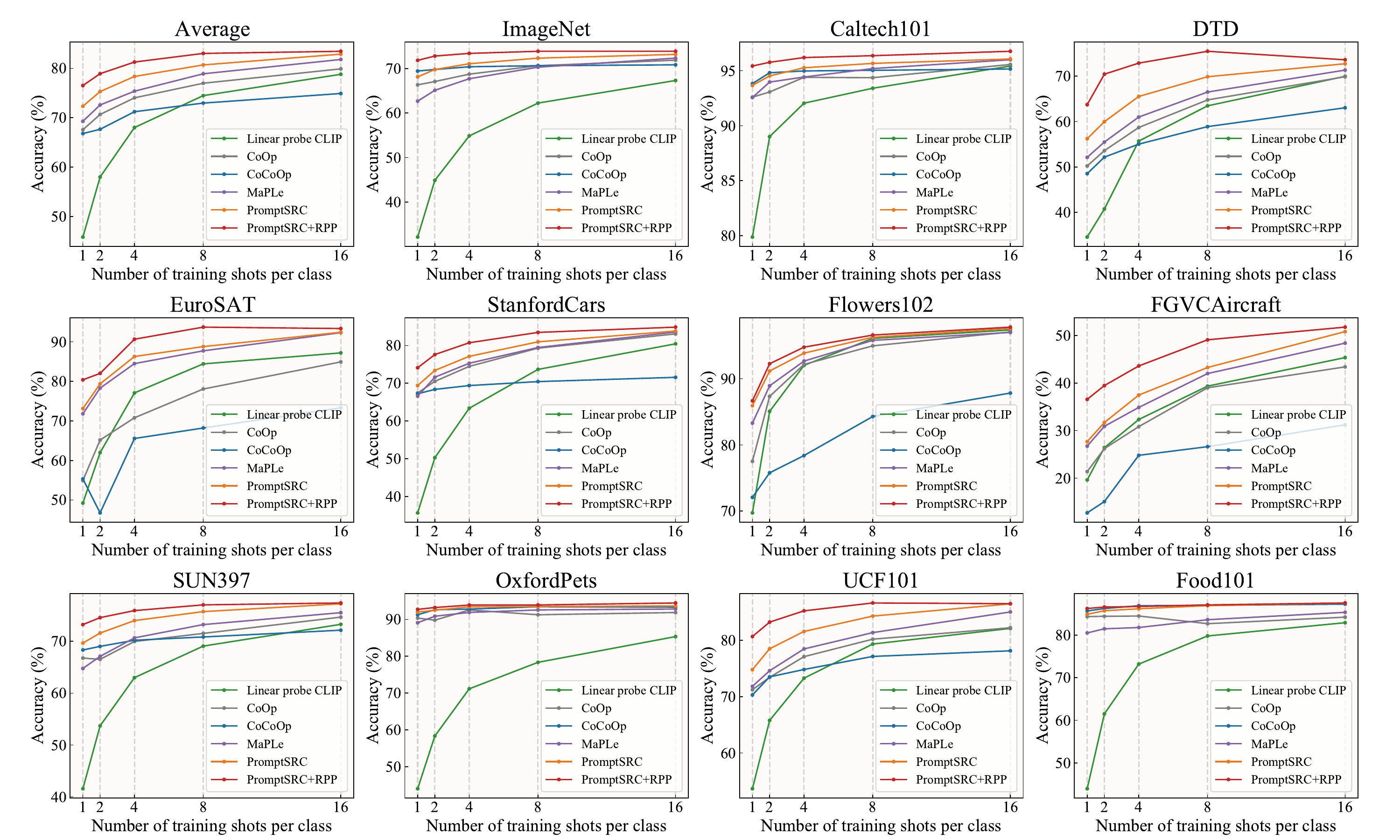}
    \vspace{-1em}
    \captionsetup{font={footnotesize}}
    \caption{\textbf{Comparison of RPP performance in a few-shot image recognition scenario.} Notably, RPP exhibits superior performance enhancement across various settings, particularly excelling in scenarios characterized by a limited number of shots.}
    \label{fig_few_shot}
    \vspace{-2em}
\end{figure}
\paragraph{\textbf{Few-shot Image Classification.}}
Tab.~\ref{table_few_shot} displays the average few-shot results of our method across eleven classification datasets. It merits particular attention that when applying RPP to other methodologies, we retain the full structure of RPP while fine-tuning only a subset of model parameters required by the downstream methods. Specifically, when employed within the PromptSRC framework, our approach, following the guidelines provided in the PromptSRC paper, exclusively fine-tunes the first nine layers of the learnable prompts, keeping the final three layers fixed. Under this fine-tuning strategy, our method achieves an enhancement of 0.58$\%$ over the existing SOTA approach represented by PromptSRC. 

Figure~\ref{fig_few_shot} illustrates the experimental results from numerous few-shot learning scenarios across eleven datasets. The red lines at the top of each graph represent our method, showcasing robust performance across all datasets under different few-shot fine-tuning conditions. Owing to an enhanced fitting pretraining while retaining the inherent generalization prowess of CLIP, particularly in data-scarce settings like 1-shot and 2-shot, our method, benefiting from comprehensive pretraining, can improve PromptSRC's average performances by 4.19$\%$ and 3.62$\%$, respectively. 
For detailed outcomes, please refer to the \textit{appendix} Sec.~\ref{Sec.D}.


\paragraph{\textbf{Base-to-New Image Classification.}}
\input{tables/base-to-new-all}

Tab.~\ref{table:base-to-novel-all} presents the base-to-new experiments of our method across eleven classification datasets.
It is noteworthy that since PromptSRC employs a strategy of incorporating learnable prompts at every layer, POMP is limited to initializing only the input token prompts with pretrained weights, while the remaining learnable parameters are subject to random initialization. This culminates in a compromise when applying it to the PromptSRC; it not only diminishes the generalized capabilities afforded by pretraining but also attenuates the fine-tuning efficacy for downstream tasks, which results in a 1.64$\%$ decline in HM.
For detailed outcomes, please refer to the \textit{appendix} Sec.~\ref{Sec.D}. 

\paragraph{\textbf{Ablation Study.}}
We assess the model's fitting ability on the validation set of ImageNet-21K and verify its generalization capacity through the average accuracy obtained from cross-dataset and cross-domain evaluations for image classification. Tab.~\ref{table:ablation-study} of \textit{appendix} Sec.~\ref{Sec.C} introduces the ablation experiments of SAPL and PPKD in detail. SAPL alone improves the model's fitting ability (21K-val +0.9$\%$), while PPKD alone improves the model's generalization ability (Zero-shot +0.2$\%$). Together, SAPL and PPKD increase 21K-val by 0.9$\%$ and Zero-shot by 0.43$\%$.

\subsection{Confirmatory Experiments}
\label{sec.ce.4.2}
\input{tables/Under_Ana}

\paragraph{\textbf{Analysis for the underfitting.}}
We consider two potential factors that may lead to underfitting: Parameter Quantity and Parameter Diversity. POMP adds unique learnable parameters only at the input layer, which results in both a smaller Parameter Quantity and lower Parameter Diversity. In contrast, our proposed SAPL is designed with a layer-by-layer replaceable QKV unshared prompt structure, which not only increases Parameter Quantity but also enriches Parameter Diversity, thereby enhancing the model's fitting capability. As shown in Tab.~\ref{table3}, we construct two comparative experiments to validate our hypothesis. The results demonstrate that limited Parameter quantity or diversity diminishes the model's fitting ability, further supporting our argument.

\paragraph{\textbf{Trade-off of learnable parameters.}}
The ablation experiment in Tab.~\ref{table3}'s Parameter Diversity illustrates that employing unshared QKV enhances fitting and generalization more effectively than merely adding shared learnable parameters. Our innovative method of enabling prompt diversification within layers through unshared QKV in prompt learning transcends a fundamental trade-off between parameter count and model performance.

\input{tables/Com_Res}

\paragraph{\textbf{Computational resource analysis.}}
Table~\ref{table2} illustrates that the RPP model requires merely an additional 166K training parameters compared to the baseline method, yet it manages to achieve similar inference flops and training durations. Moreover, it demonstrates effective performance across a range of tasks. This suggests that the RPP model strikes an efficient balance between computational efficiency and strong performance.

\section{Conclusion}
\label{sec.5}
We are at the forefront of recognizing and tackling underfitting issues during Prompt Pretraining in VLMs, with a focus on refining prompt structure and supervision. Theoretical analysis corroborates the enhanced generalization capability of our method. Our approach signifies a paradigm shift in prompt learning research, emphasizing the importance of robust initialization strategies. Nevertheless, two challenges persist: 1) Insufficient exploration of the hierarchical category data available in ImageNet-21K. 2) Constraints imposed by the form of supervised pretraining and label limitations impede the acquisition of generalized knowledge. These aspects warrant further investigation in future research endeavors.

\section*{Acknowledgement}
This research was supported by the Young Scientists Fund of the National Natural Science Foundation of China (Grant No.62206134), the Fundamental Research Funds for the Central Universities 070-63233084, and the Tianjin Key Laboratory of Visual Computing and Intelligent Perception (VCIP). Computation is supported by the Supercomputing Center of Nankai University (NKSC).

\bibliographystyle{plain}
\bibliography{neurips_2024}
\newpage
\renewcommand{\thetable}{S\arabic{table}}
\renewcommand{\thefigure}{S\arabic{figure}}
\renewcommand{\theequation}{S\arabic{equation}}
\setcounter{equation}{0}
\setcounter{table}{0}
\setcounter{figure}{0}
\appendix
\section*{Appendix} 
In the appendix, we will provide detailed descriptions of formula proofs, parameter settings for all experiments, datasets, and evaluation metrics. We will include comprehensive ablation studies and comparative experiments, and finally, analyze the broader impacts and safeguards.



\section{Theoretical Proof}
\label{sec.A}
This section provides detailed proofs for the Theorem in Sec.~3.5. We introduce the following lemmas for proving our Theorem. 

\textbf{Lemma 1}(McDiarmid's Inequality~\cite{vershynin2018high})\textbf{.} \textit{Consider independent random variables $v_{1},v_{2},\cdots ,v_{n}\in \mathcal{V}$ and a function $\phi:\mathcal{V}^{n}\to \mathbb{R}$. Suppose that for all $v_{1},v_{2},\cdots ,v_{n}$ and ${v_{i}}'\in \mathcal{V}\left ( i=1,2,\cdots ,n \right ) $, the function satisfies}
\begin{equation}
    \left | \phi\left ( v_{1},\cdots ,V_{i-1},V_{i},V_{i+1},\cdots,V_{n} \right ) -\phi\left ( v_{1},\cdots ,V_{i-1},{v_{i}}',V_{i+1},\cdots,V_{n} \right )  \right | \le c_{i},
    \vspace{0pt}
    \label{lemma1_1}
\end{equation}
\textit{and then it holds that}
\begin{equation}
    \mathcal{P}\left \{ \phi\left ( v_{1},v_{2},\cdots,v_{n} \right ) -\mathbb{E}_{v_{1},v_{2},\cdots,v_{n}}\left (  \phi\left ( v_{1},v_{2},\cdots,v_{n} \right )\right ) > \mu  \right \}  \le e^{-\frac{2\mu ^{2}}{ {\textstyle \sum_{i=1}^{n}c_{i}^{2}} } }.
    \vspace{0pt}
    \label{lemma1_2}
\end{equation}

\textbf{Lemma 2.} \textit{Let $\Theta^{*} $ be the solution to the optimization objective}
\begin{equation}
    \Theta^{*} \in \underset{\Theta \in \mathbb{R}^{d}}{\mathrm{min}}\frac{1}{N} {\textstyle \sum_{i=1}^{N}}\mathcal{L}\left ( \hat{s}_{i}^{S}\left ( \Theta \right )  ,y_{i}^{gt} \right ) +\lambda\frac{1}{N\cdot K}  {\textstyle \sum_{k=1}^{K}} {\textstyle \sum_{i=1}^{N}}\hat{s}_{ik}^{T}\mathrm{log}\frac{\hat{s}_{ik}^{T}}{\hat{s}_{ik}^{S}\left ( \Theta \right )},
    \vspace{0pt}
    \label{lemma2_1}
\end{equation}
\textit{then there exists a bounded tensor set $\mathcal{F}\left ( \lambda \right ) $ such that}
\begin{equation}
    \Theta ^{*}\in \mathcal{F}\left ( \lambda \right ) =\left \{ \Theta |\: e^{\frac{C_{0}}{\lambda s_{ik}^{T}}} \le s_{ik}^{S}\left ( \Theta \right ) \le 1,i\in \mathbb{N}_{N}, k\in \mathbb{N}_{K} \right \},
    \vspace{0pt}
    \label{lemma2_2}
\end{equation}
\textit{where the constant $C_{0} >0$ is not dependent on $\lambda$.}

\begin{proof}
    According to the optimality of $\Theta ^{*}$, it follows that
    \begin{equation}
        \begin{aligned}
            &\frac{1}{N} {\textstyle \sum_{i=1}^{N}}  \mathcal{L}\left ( \hat{s}_{i}^{S}\left ( \Theta^{*} \right )  ,y_{i}^{gt} \right )+\lambda\frac{1}{N\cdot K}  {\textstyle \sum_{k=1}^{K}} {\textstyle \sum_{i=1}^{N}}\hat{s}_{ik}^{T}\mathrm{ log}\frac{\hat{s}_{ik}^{T}}{\hat{s}_{ik}^{S}\left ( \Theta \right )} \\
            &\le \frac{1}{N} {\textstyle \sum_{i=1}^{N}}  \mathcal{L}\left ( \hat{s}_{i}^{T}  ,y_{i}^{gt} \right ) +\lambda\frac{1}{N\cdot K}  {\textstyle \sum_{k=1}^{K}} {\textstyle \sum_{i=1}^{N}}\hat{s}_{ik}^{T}\mathrm{log}\frac{\hat{s}_{ik}^{T}}{\hat{s}_{ik}^{T}} \\
            &\le \frac{1}{N} {\textstyle \sum_{i=1}^{N}}  \mathcal{L}\left ( \hat{s}_{i}^{T}  ,y_{i}^{gt} \right ).
            \vspace{0pt}
            \label{lemma2_2}
        \end{aligned}
    \end{equation}

    We denote that $\mathcal{L}_{min}= \underset{\Theta^{*},i=1,2,\cdots,N}{\mathrm{inf} } \mathcal{L}\left ( \hat{s}_{i}^{S}\left ( \Theta^{*} \right )  ,y_{i}^{gt} \right )$, and have that
    \begin{equation}
        \begin{aligned}
            &\lambda\frac{1}{N\cdot K}  {\textstyle \sum_{k=1}^{K}} {\textstyle \sum_{i=1}^{N}}\hat{s}_{ik}^{T}\mathrm{log}\frac{\hat{s}_{ik}^{T}}{\hat{s}_{ik}^{S}\left ( \Theta \right )}\\
            &\Leftrightarrow  -\lambda\frac{1}{N\cdot K}  {\textstyle \sum_{k=1}^{K}} {\textstyle \sum_{i=1}^{N}}\hat{s}_{ik}^{T}\mathrm{log}\hat{s}_{ik}^{S}\left ( \Theta \right )\\
            &\le \frac{1}{N} {\textstyle \sum_{i=1}^{N}}  \mathcal{L}\left ( \hat{s}_{i}^{T}  ,y_{i}^{gt} \right )-\frac{1}{N} {\textstyle \sum_{i=1}^{N}}  \mathcal{L}\left ( \hat{s}_{i}^{S}\left ( \Theta^{*} \right )  ,y_{i}^{gt} \right )\\
            &\le \frac{1}{N} {\textstyle \sum_{i=1}^{N}}  \mathcal{L}\left ( \hat{s}_{i}^{T}  ,y_{i}^{gt} \right )-\frac{1}{N} {\textstyle \sum_{i=1}^{N}}\mathcal{L}_{min}\\
            &=C_{0},
            \vspace{0pt}
            \label{lemma2_3}
        \end{aligned}
    \end{equation}
    where $C_{0}>0$. Finally, we have
    \begin{equation}
        e^{\frac{C_{0}}{\lambda s_{ik}^{T}}} \le s_{ik}^{S}\left ( \Theta \right ) \le 1,
        \vspace{0pt}
        \label{lemma2_4}
    \end{equation}   
    which completes the proof.
\end{proof}

The proof of Theorem~1. is given as follows.

\textbf{Theorem 1.} \textit{Assume that $\Theta ^{*}$ is the solution to RPP. Then we have that for any $0<\delta <1$ with probability $1-\delta $,} 
\begin{equation}
    \varepsilon\left ( \Theta^{*}  \right )-\bar{\varepsilon} _{\chi }\left ( \Theta^{*}   \right )\le X^{*}\sqrt{2ln\left ( 1/\delta  \right )/N }+B_{\lambda }R_{N}\left ( \mathcal{L} \right ) , 
    \vspace{0pt}
    \label{theorem2_1}
\end{equation}
 \textit{where $B_{\lambda}\to 0$ as $\lambda\to +\infty$, and $X^{*}=\mathrm{max}_{r\in \mathbb{N}_{N}}\left |  \mathcal{L}\left( \hat{s}_{r}^{s}\left( \Theta \right), y_{r}^{gt} \right) \right | $. Here $R_{N}\left ( \mathcal{L} \right )$ is the Rademacher complexity of the loss function $\mathcal{L}$ related to the space $\mathbb{R}$ for $N$ training examples.} 

\begin{proof}
    Firstly, we denote that
    \begin{equation}
        \varepsilon \left ( \Theta^{*} \right )=\frac{1}{N} {\textstyle \sum_{i=1}^{N}}  \mathcal{L}\left ( \hat{s}_{i}^{S}\left ( \Theta \right )  ,y_{i}^{gt} \right ),
        \vspace{0pt}
        \label{theorem2_2}
    \end{equation}
    and 
    \begin{equation}
        \bar{\varepsilon}_{\chi, r} \left( \Theta^{*} \right) = \frac{1}{N} \left( \sum _{i=1,i\ne r}^{N} \mathcal{L}\left( \hat{s}_{i}^{S}\left( \Theta \right), y_{i}^{gt} \right) + \mathcal{L}\left( \hat{a}_{r}^{S}\left( \Theta \right), b_{r}^{gt} \right) \right),
        \label{theorem2_3}
    \end{equation}
    where $\left ( \hat{a}_{k}^{S}\left ( \Theta \right ),b_{k}^{gt} \right ) $ is an arbitrary data pair from the sample space with similarity label $b_{k}^{gt}$. Then we have that
    \begin{equation}
        \begin{aligned}
        \left | \varepsilon \left ( \Theta^{*} \right )-\bar{\varepsilon}_{\chi,r} \left( \Theta^{*} \right) \right | &= \frac{1}{N}\left |  \mathcal{L}\left( \hat{s}_{r}^{S}\left( \Theta \right), y_{r}^{gt} \right) -\mathcal{L}\left( \hat{a}_{r}^{S}\left( \Theta \right), b_{r}^{gt} \right)\right |\\
        &\le \frac{1}{N}\left ( \left |  \mathcal{L}\left( \hat{s}_{r}^{S}\left( \Theta \right), y_{r}^{gt} \right)\right |+\left | \mathcal{L}\left( \hat{a}_{r}^{S}\left( \Theta \right), b_{r}^{gt} \right) \right |  \right )\\
        &\le \frac{2}{N}X^{*},
        \end{aligned}
        \label{theorem2_4}
    \end{equation}
    where $X^{*}=\mathrm{max}_{r\in \mathbb{N}_{N}}\left |  \mathcal{L}\left( \hat{s}_{r}^{S}\left( \Theta \right), y_{r}^{gt} \right) \right | $. Then we apply Lemma 2 to the term $\varepsilon \left ( \Theta^{*} \right )-\bar{\varepsilon}_{\chi} \left( \Theta^{*} \right)$ and have that with probability $1-\delta $ it holds that 
    \begin{equation}
    \varepsilon \left ( \Theta^{*} \right )-\bar{\varepsilon}_{\chi} \left( \Theta^{*} \right)\le \mathbb{E}_{\chi}\left ( \varepsilon \left ( \Theta^{*} \right )-\bar{\varepsilon}_{\chi} \left( \Theta^{*} \right) \right ) + X^{*}\sqrt{2ln\left ( 1/\delta  \right )/N }.
    \label{theorem2_5}
    \end{equation}

    Now we only need to estimate the first term of the right-hand side of the above inequality. Specifically, there holds
    \begin{equation}
        \mathbb{E}_{\chi}\left ( \varepsilon \left ( \Theta^{*} \right )-\bar{\varepsilon}_{\chi} \left( \Theta^{*} \right) \right )=\mathbb{E}_{\chi}\left (\mathbb{E}_{\mathcal{Z}} \varepsilon_{\mathcal{Z}} \left ( \Theta^{*} \right )-\bar{\varepsilon}_{\chi} \left( \Theta^{*} \right) \right )\le\mathbb{E}_{\chi,\mathcal{Z}}\left ( \varepsilon_{\mathcal{Z}} \left ( \Theta^{*} \right )-\bar{\varepsilon}_{\chi} \left( \Theta^{*} \right) \right ),
    \label{theorem2_6}
    \end{equation}
    where $\mathcal{Z}=\left \{ z_{i} | z_{i}\sim \mathcal{D},i\in \mathbb{N}_{N} \right \} $ are independent identically distributed (i.i.d.) samples which are independent of $\mathcal{\chi}=\left \{ x_{i} | x_{i}\sim \mathcal{D},i\in \mathbb{N}_{N} \right \} $. By Lemma 2, we know that there exists the bounded tensor set $\mathcal{F}\left ( \lambda \right ) $ such that 
    \begin{equation}
        \Theta ^{*}\in \mathcal{F}\left ( \lambda \right ) =\left \{ \Theta| e^{\frac{C_{0}}{\lambda s_{ik}^{T}}} \le s_{ik}^{S}\left ( \Theta \right ) \le 1,i\in \mathbb{N}_{N}, k\in \mathbb{N}_{K} \right \},
        \vspace{0pt}
        \label{theorem2_7}
    \end{equation}    
    where $C_{0}>0$ is a constant. Let the function
    \begin{equation}
         B_{\lambda }=2\mathbb{E}_{\chi,\mathcal{Z}}\left ( \underset{\Theta\in \mathcal{F}\left ( \lambda \right )}{\mathrm{sup}}  \bar{\varepsilon}_{\mathcal{Z}} \left( \Theta \right)-\bar{\varepsilon}_{\chi} \left( \Theta \right) \right )/\mathbb{E}_{\chi,\mathcal{Z}}\left ( \underset{\Theta\in \mathbb{R}}{\mathrm{sup}}  \bar{\varepsilon}_{\mathcal{Z}} \left( \Theta \right)-\bar{\varepsilon}_{\chi} \left( \Theta \right) \right ).
        \vspace{0pt}
        \label{theorem2_7}
    \end{equation}
    
    By \textit{Levi’s Monotone Convergence Theorem}~\cite{krantz2002primer}, we have
    \begin{equation}
        \begin{aligned}
        &\lim_{\lambda  \to \infty} \mathbb{E}_{\chi,\mathcal{Z}}\left ( \underset{\Theta\in \mathcal{F}\left ( \lambda \right )}{\mathrm{sup}}  \bar{\varepsilon}_{\mathcal{Z}} \left( \Theta \right)-\bar{\varepsilon}_{\chi} \left( \Theta \right) \right ) \\
        &=\mathbb{E}_{\chi,\mathcal{Z}}\left ( \lim_{\lambda  \to \infty}\underset{\Theta\in \mathcal{F}\left ( \lambda \right )}{\mathrm{sup}}  \bar{\varepsilon}_{\mathcal{Z}} \left( \Theta \right)-\underset{\Theta\in \mathcal{F}\left ( \lambda \right )}{\mathrm{sup}} \bar{\varepsilon}_{\chi} \left( \Theta \right) \right )\\
        &=\mathbb{E}_{\chi,\mathcal{Z}}\left (  \bar{\varepsilon}_{\mathcal{Z}}\left ( 0 \right ) - \bar{\varepsilon}_{\mathcal{\chi}}\left ( 0 \right )  \right )\\
        &= \mathbb{E}_{\mathcal{Z}}\left (  \bar{\varepsilon}_{\mathcal{Z}}\left ( 0 \right ) \right )-\mathbb{E}_{\chi,}\left (   \bar{\varepsilon}_{\mathcal{\chi}}\left ( 0 \right )\right ) \\
        &=0.
        \end{aligned}
        \label{theorem2_8}
    \end{equation}

    Therefore, we obtain $\lim_{\lambda  \to \infty} B_{\lambda} =0.$ By standard symmetrization techniques for i.i.d. Rademacher variables $\sigma =\left ( \sigma_{1},\sigma_{2},\cdots,\sigma_{N} \right )^{\top }$, it follows that
     \begin{equation}
        \begin{aligned}
        \mathbb{E}_{\chi,\mathcal{Z}}\left (  \bar{\varepsilon}_{\mathcal{Z}} \left ( \Theta^{*} \right )-\bar{\varepsilon}_{\chi} \left( \Theta^{*} \right) \right )&\le\mathbb{E}_{\chi,\mathcal{Z}}\left ( \underset{\Theta\in \mathcal{F}\left ( \lambda \right )}{\mathrm{sup}}  \bar{\varepsilon}_{\mathcal{Z}} \left ( \Theta \right )-\bar{\varepsilon}_{\chi} \left( \Theta \right) \right )  \\
        &=\frac{B_{\lambda }}{2} \mathbb{E}_{\chi,\mathcal{Z}}\left ( \underset{\Theta\in \mathbb{R}}{\mathrm{sup}}\bar{\varepsilon}_{\mathcal{Z}} \left ( \Theta \right )-\bar{\varepsilon}_{\chi} \left( \Theta \right)  \right ) \\
        &=\frac{B_{\lambda }}{2N} \mathbb{E}_{\chi,\mathcal{Z},\sigma }\left ( \underset{\Theta^{*}\in \mathbb{R}}{\mathrm{sup}} {\textstyle \sum_{i=1}^{N}}\sigma _{i}\left ( \mathcal{L}\left ( s_{i}^{S}\left ( \Theta  \right )  \right )-  \mathcal{L}\left ( z_{i}^{S}\left ( \Theta  \right )  \right )\right )   \right ) \\
        &=\frac{B_{\lambda }}{N} \mathbb{E}_{\chi,\sigma } \left ( \underset{\Theta^{*}\in \mathbb{R}}{\mathrm{sup}} {\textstyle \sum_{i=1}^{N}}\sigma_{i} \mathcal{L}\left ( s_{i}^{S}\left ( \Theta  \right )  \right ) \right ) \\
        &=B_{\lambda }R_{N}\left ( \mathcal{L} \right ),
        \end{aligned}
        \label{theorem2_9}
    \end{equation}
    where  $P\left \{ \sigma _{i}=0 \right \} =P\left \{ \sigma _{i}=1 \right \} =0.5$ for $i\in \mathbb{N}_{N}$, and $R_{N}\left ( \mathcal{L} \right )$ is the Rademacher complexity of L. Finally, combining the above inequality with Eqn.~\eqref{theorem2_5} and Eqn.~\eqref{theorem2_6} completes the proof.
\end{proof}

\section{Training details}
\label{Sec.B}
\subsection{Text templates and Datasets}
\label{secB.2}
\input{tables/templates}
\paragraph{\textbf{Text templates for senior teacher CLIP}}
According to prior research~\cite{khattak2023self}, different prompt templates are employed for various datasets to enhance the text representation capability of the senior CLIP and to improve the distillation effect of the boosting prompts. Specifically, the corresponding template for each dataset is shown in Tab.~\ref{tab:dataset}.

\paragraph{\textbf{Prompt Pretraining.}} 
Following the conventional ImageNet-21K pretraining approach~\cite{ridnik2021imagenet}, we undertake the following three processes on the dataset: (1) Invalid class filtering: To mitigate the influence of an extremely long-tail distribution on experimental outcomes, classes with fewer than 500 images are excluded. Consequently, from the fall 11 release, the dataset comprises 12,358,688 images spanning 11,221 classes. (It's noteworthy that POMP utilizes the winter 21 version, necessitating all comparison experiments with POMP to be replicated on our dataset). (2) Creation of a validation set: For standardization and future benchmarking purposes, we allocate 50 images per class for a uniform validation split. (3) Image resizing: To facilitate accessibility and expedite training, all images within the ImageNet-21K dataset are resized to a resolution of 224 during the preprocessing phase. All methods train on the full ImageNet-21K dataset (fall 11 version). The specifics of the pretraining dataset, ImageNet-21K, utilized for Prompt Pretraining, are presented in Tab.~\ref{tab:dataset}.

It's imperative to note that the results shown are replicated on our proprietary dataset due to discrepancies with the dataset employed by POMP. Furthermore, our advocated SAPL prompt structure necessitates tailored modifications to align with the ViT architecture. Consequently, when employing ResNet as the backbone, this innovative structure is exclusively integrated within the text encoder.

\paragraph{\textbf{Image Classification.}} 
For zero/few-shot and base-to-new image classification, we evaluate the performance of PPKD on $11$ downstream datasets, including Caltech-101, Oxford-Pets, Stanford Cars, Oxford-Flowers102, Food-101, FGVC Aircraft, EuroSAT, SUN-397, Describable Textures (DTD), UCF-101, ImageNet. We also conduct zero-shot evaluation on 3 out-of-domain datasets including ImageNetV2 ImageNet-S, and ImageNet-R, to evaluate the domain generalization capability of our method. The specifics of the downstream datasets utilized for classification are presented in Tab.~\ref{tab:dataset}.

\subsection{Task Setups and Implementation Details}
\label{secB.1}
\subsubsection{Pretraining Details}
Our experiments are conducted utilizing 8$\times$Nvidia A40 GPUs. For the pretraining phase, we employ the proposed ImageNet-21K fall 11 version~\cite{ridnik2021imagenet}. Following POMP~\cite{ren2023prompt}, each class is represented by 16 training samples (16-shot), and the prompt length is set to 4. The batch size used is 32, and the maximum epoch is limited to 5. At each training step, we sample 1,000 classes, denoted as K=1000. We employ the SGD optimizer with an initial learning rate (lr) of 0.016, decay according to the cosine annealing rule.

\subsubsection{Setting for Image Classification}
Our experiments are conducted utilizing 8$\times$Nvidia A40 GPUs. We perform an extensive comparative experimental analysis of our proposed method in comparison with the prior SOTA POMP approach. This evaluation involves utilizing the PromptSRC~\cite{khattak2023self} and CoOp\cite{zhou2022learning} frameworks within the base-to-new experimental setting. 

In the few-shot experiments, we utilize a standardized experimental setup, maintaining a fixed batch size of $4$ and conducting a maximum of $50$ epochs. Specifically within the PromptSRC framework, the lr is set at $0.02$. However, in the base-to-new experiments, we reduce the maximum number of epochs from $50$ to $20$. In the PromptSRC framework, the lr remained at $0.02$, while it is adjusted to $0.036$ within the CoOp framework.

Our novel prompt structure necessitates a tailored approach to model weight initialization, which may involve fine-tuning to achieve compatibility with established training protocols or selectively merging new structural elements with extant pretrained weights.

\section{Ablation study}
\label{Sec.C}
\input{tables/ablation-study}
We initially conduct ablation studies focusing on two aspects: prompt structure (SAPL) and prompt supervision (PPKD). According to the data presented in the second and third rows of Tab.~\ref{table:ablation-study}, it is evident that both SAPL and PPKD methodologies are capable of enhancing the pretraining model's fitting ability, as reflected by an increase in Pretrained validation accuracy. However, in alignment with the analysis discussed in the main body, excessive fitting may lead to a diminishment in the model's generalization capabilities inherited from CLIP, as indicated by a \descend{1.46$\%$} decrease in zero-shot performance reported in the second row. To mitigate this issue, we additionally utilize soft labels derived from zero-shot probability predictions provided by a large-scale CLIP teacher model. As shown in the fourth row, when SAPL and PPKD are employed in conjunction, there is an observed improvement of \improve{1.2$\%$} in our pretraining validation set accuracy, along with a \improve{0.21 $\%$} augmentation in the average precision of zero-shot tasks.

Subsequently, according to the ``Ablation Study" of Tab.~\ref{table:ablation-study}, we undertake ablation experiments related to hyperparameters. Based on these results, we discern that indiscriminately augmenting model parameters results in a profound loss of generalization. Consequently, we ascertain the ideal parameter configuration for the model by equilibrating the degree of pretraining with the capacity for downstream generalization, as illustrated by the \colorbox{tablecolor}{gray background}. It is worth noting that in ablation experiments, our more selective criterion is the generalization ability of the model under certain fitting ability.

\section{Additional experiments}
\label{Sec.D}
\paragraph{\textbf{Base-to-New}}
\input{tables/base-to-new}
\input{tables/few-shot-allshot}
Tab.~\ref{table:base-to-novel} presents the empirical results of the base-to-new generalization experiments comparing our approach with existing methods. Our method demonstrates an average improvement of \improve{7.73$\%$} and \improve{1.13$\%$} over the baseline methods CoOp and PromptSRC, respectively, across average on eleven datasets. Compared to the prior pretrained method POMP, improvements are observed at \improve{5.3$\%$} and \improve{2.77$\%$}, respectively. Notably, in datasets where CLIP's generalization performance is suboptimal, such as FGVCAircraft and DTD datasets, the use of POMP as a baseline results in a notable decrease in Harmonic Mean (HM) by \descend{9.11$\%$} and \descend{8.69$\%$}, respectively. This decline is attributable to POMP's underfitting during pretraining on ImageNet-21K and the subsequent loss of CLIP’s inherent generalization capabilities, leading to a narrower convergence direction when its pretrained weights are employed for initialization. In contrast, our method, benefiting from both the fitting ability of the pretraining and the original CLIP's generalizability, manages to achieve performance gains of \improve{0.63$\%$} and \improve{0.9$\%$} over PromptSRC.

\paragraph{\textbf{Few-shot}}
Tab.~\ref{tab_appendix:few_shot_experiments} presents the fine-tuning experimental results of our method compared with existing approaches across a spectrum of shots. Owing to the appropriate fit of pretraining and retention of the inherent generalization capabilities of CLIP, our method yields an exceptionally notable enhancement in performance when training with a minimal number of data samples (1-shot, 2-shot), relative to existing methods. This is particularly evident on challenging datasets such as FGVCAircraft, where our method delivers performance improvements of \improve{8.9$\%$} and \improve{7.77$\%$} over PromptSRC at 1-shot and 2-shot, respectively. These findings further substantiate the value of the pretraining strategy we propose, termed RPP.

\section*{Broader Impacts}
Further research and careful consideration are necessary when utilizing this technology, as the presented proposed method relies on statistics derived from training datasets that may possess biases and could potentially result in negative societal impacts.

\section*{Safeguards}
Our paper employs the ImageNet-21K dataset for pretraining in an open-source multimodal model. Potential security concerns may arise from biases in the pretraining of open-source data and multimodal models. Please be mindful of biases in the original data and model, as well as the security of the model. We do not release any data or models; we only provide a pretraining approach.

\end{document}

%% file: tables/pretrain.tex
\begin{wraptable}{rt}{0.5\linewidth}
    \centering
    \captionsetup{font={scriptsize}}
    \vspace{-11.5pt}
    \captionof{table}{
	\textbf{Performance on the ImageNet-21K validation set.} The backbones of our experiments are ResNet-50 (teacher: ResNet-101) and ViT/B-16 (teacher: ViT/L-14). In the upper tier, ZeroshotCLIP and Prompt Ensemble implement zero-shot inference. ``-" indicates that the item is empty. ``our impl." means our implementation of these methods.}
	\label{pretrain-table}
    \resizebox{0.99\linewidth}{!}{
	\begin{tabular}{lcc}
    \toprule
    Method    & ResNet50    & ViT-B/16      \\ \midrule
    ZeroshotCLIP~\cite{radford2021learning}  & 17 & 20.7          \\ 
    Prompt Ensemble~\cite{radford2021learning}  & 18.8 & 23.5          \\  
    Linear Probing (our impl.)~\cite{radford2021learning}    & 5.9    & 20.3          \\
    \midrule
    VPT (our impl.)~\cite{Derakhshani2022VariationalPT}        &  -     & 23.6          \\
    POMP (our impl.)~\cite{ren2023prompt}    & 19.4 & 24 \\ 
    \rowcolor{tablecolor} \textbf{RPP (ours)}  & \textbf{20.2}\improve{(+0.8)} & \textbf{24.9}\improve{(+0.9)} \\
    \bottomrule
    \end{tabular}
	\vspace{-2em}
    }
\end{wraptable}

%% file: tables/zero-shot.tex
\begin{table}[]
\captionsetup{font={scriptsize}}
\caption{\textbf{Cross-dataset and cross-domain evaluation for image classification.} The backbone is ViT/B-16. Overall, RPP secures the highest mean accuracy, denoting superior generalization capabilities. The designation ``–" signifies that the corresponding entry is unpopulated or not applicable. Each number in the figure represents the validation set accuracy (\%) of the current dataset.}
\begin{adjustbox}{max width=\textwidth}
\begin{tabular}{l c cc lllllllllll lllll}
\toprule
&\multicolumn{2}{c}{\textbf{Source}}& \multicolumn{11}{c}{\textbf{Target (cross-dataset)}} & \multicolumn{5}{c}{\textbf{Target (cross-domain)}} \\ \addlinespace[4pt] \cmidrule(lr){2-3}\cmidrule(lr){4-14} \cmidrule(lr){15-19} \addlinespace
&\rotbox{ImageNet-21K} &\rotbox{ImageNet} & \rotbox{Caltech101} & \rotbox{OxfordPets} & \rotbox{StanfordCars} & \rotbox{Flowers102} & \rotbox{Food101} & \rotbox{FGVCAircraft} & \rotbox{SUN397} & \rotbox{DTD} & \rotbox{EuroSAT} & \rotbox{UCF101} & \rotbox{\textbf{\emph{Average}}} & \rotbox{ImageNetV2} & \rotbox{ImageNet-S} & \rotbox{ImageNet} & \rotbox{ImageNet-R} & \rotbox{\textbf{\emph{Average}}} \\
\midrule
CoOp~\cite{zhou2022learning} &-&\checkmark & 93.7 & 89.1 & 64.5 & 68.7 & 85.3 & 18.5 & 64.2 & 41.9 & 46.4 & 66.6 & 63.9 & 64.2 & 48.0  &-& 75.2 & 62.5 \\
CoCoOp~\cite{zhou2022conditional}&-&\checkmark & 94.4 & 90.1 & 65.3 & \textbf{71.9} & 86.1 & 22.9 & 67.4 & 45.7 & 45.4 & 68.2 & 65.7 & 64.1 & 48.8 &-& 76.2 & 63.0  \\
VPT~\cite{Derakhshani2022VariationalPT}&-&\checkmark & 93.7 & 90.6 & 65.0 & 70.9 & \textbf{86.3} & \textbf{24.9} & 67.5 & 46.1 & 45.9 & 68.7 & 66.0  & 64.2 & 49.2 &-& 77.0 & 63.5 \\
PromptSRC~\cite{khattak2023self}&-&\checkmark & 93.6 & 90.3 & 65.7 & 70.3 & 86.2 & 24.0 & 67.1 & \textbf{46.9} & 45.5 & \textbf{68.8} & 65.8 & \textbf{64.4} & \textbf{49.6} &-& 77.8 & 63.9 \\
\midrule
hard prompt &- &-& 93.3 & 88.2 & 65.6 & 67.4 & 85.3 & 23.7 & 62.6 & 44.3 & 42.0& 65.1 & 63.7 & 60.9 & 46.1 &66.7 & 74.0 & 61.9 & \\ 
POMP (our impl.)~\cite{ren2023prompt}&\checkmark &-& 94.5 & \textbf{90.9} & 65.5 & 71.7 & 86.1 & 23.9 & 66.9 & 44.3 & 47.3 & 66.8 & 65.8 & 63.4 & 48.8 & 69.9  & 76.9 & 64.7  \\ 
\rowcolor{tablecolor} \textbf{RPP (ours)}&\checkmark &-& \textbf{94.5} & 90.0 & \textbf{66.3} & 71.4 & 85.8 & 23.1 & \textbf{68.3} & 45.4 & \textbf{47.9} & 68.4 & \textbf{66.1} & 63.7 & 49.5 & \textbf{70.7}  & \textbf{78.0} & \textbf{65.5}  \\ 
\bottomrule
\label{zero-shot}
\vspace{-2em}
\end{tabular}
\end{adjustbox}
\end{table}

%% file: tables/few-shot.tex
\begin{wraptable}{rt}{0.5\linewidth}
    \centering
    \captionsetup{font={scriptsize}}
    \vspace{-12.5pt}
    \captionof{table}{
	\textbf{Few-shot experiment over 11 image classification datasets,} focusing on the average results of all datasets. The backbone is ViT/B-16. For various shot results of each dataset, see Fig.~\ref{fig_few_shot}. }
	\label{table_few_shot}
    \resizebox{0.99\linewidth}{!}{
	\begin{tabular}{@{}lc@{}}
    \toprule
    Method & validation accuracy (\%) \\ \midrule
    Linear Probing~\cite{radford2021learning} & 78.79 \\
    CoCoOp~\cite{zhou2022conditional} &74.90 \\
    CoOp~\cite{zhou2022learning} &79.89 \\
    MaPLe ~\cite{khattak2023maple} &81.79 \\
    \midrule
    PromptSRC~\cite{khattak2023self} & 82.87  \\
    + POMP (our impl.)~\cite{ren2023prompt} & 82.56 \descend{(-0.31)} \\
    \rowcolor{tablecolor} \textbf{+ RPP (ours)} & 83.45  \improve{(+0.58)} \\
    \bottomrule
    \end{tabular}
	\vspace{-11pt}
    }
\end{wraptable}

%% file: tables/base-to-new-all.tex
\begin{wraptable}{rt}{0.5\linewidth}
    \centering
    \captionsetup{font={scriptsize}}
    \captionof{table}{
	\textbf{Comparison with state-of-the-art methods \textit{w/} or \textit{w/o} RPP on base-to-novel generalization.} The backbone is ViT/B-16. The ``base-to-new'' experiments split datasets into base and novel class sets, training on the base and testing on both to assess the model's generalization to less-represented subdomains.}
	\label{table:base-to-novel-all}
    \resizebox{0.99\linewidth}{!}{
    \begin{tabular}{lccl}
        \toprule
        Method    & Base (\%)  & Novel (\%) & HM (\%) \\
        \midrule
        CLIP~\cite{radford2021learning} & 69.34 & 74.22 & 71.70 \\
        CoCoOp~\cite{zhou2022conditional} & 80.47 & 71.69 & 75.83 \\
        MaPLe~\cite{khattak2023maple}   & 82.28 & 75.14 & 78.55 \\
        \midrule
        CoOp~\cite{zhou2022conditional} & 82.69 & 63.22 & 71.66 \\
        + POMP (our impl.) &78.51 	&70.14 	&74.09    \improve{(+2.43)} \\
        \rowcolor{tablecolor} \textbf{+ RPP (ours)} &82.19 	&76.77	&79.39    \improve{(+7.73)} \\
        \midrule
        PromptSRC~\cite{khattak2023self} & 84.26 & 76.10 & 79.97 \\
        + POMP (our impl.)~\cite{ren2023prompt} &83.48 	&73.78 	&78.33    \descend{(-1.64)} \\
        \rowcolor{tablecolor} \textbf{+ RPP (ours)} &85.21  &77.38 &81.10  \improve{(+1.13)}   \\
        \bottomrule
    \end{tabular}
	\vspace{-9pt}
    }
    \vspace{-1em}
\end{wraptable}

%% file: tables/Under_Ana.tex
\vspace{-1.5em}
\begin{table}[h!]
\captionsetup{font={footnotesize}}
\caption{\textbf{Quantitative experiment on underfitting.} SAPL\_shared denotes a layer-by-layer replaceable shared QKV. To maintain parity in parameter count between configurations employing shared and unshared QKV, we adjust the number of learnable tokens per layer for the shared QKV to be 3 times that of the unshared QKV.}
\begin{center}
\begin{adjustbox}{max width=\textwidth}
\begin{tabular}{lllccc}
\toprule
Cause of underfitting & Method  & Difference  & 21K-train (\%) & 21K-val (\%)  &Zero-shot (\%)   \\ 
\midrule
\multirow{2}{*}{Parameter Quantity} 
&RPP (PPKD) &Learning params (1-layer) & 27.0  &24.1 & 65.64 \\
&RPP (SAPL + PPKD) &Learning params (12-layer)  &29.3   &24.9 &65.92 \\ 
\midrule
\multirow{2}{*}{Parameter Diversity} 
&RPP (SAPL\_shared + PPKD) &Shared QKV  &26.0   &23.3 &64.82 \\
&RPP (SAPL + PPKD) &Unshared QKV  &29.3   &24.9 &65.92\\ 
\bottomrule
\end{tabular}
\end{adjustbox}
\end{center}
\label{table3}
\vspace{-1.5em}
\end{table}

%% file: tables/Com_Res.tex
\begin{wraptable}{rt}{0.6\linewidth}
    \centering
    \captionsetup{font={scriptsize}}
    \vspace{-12.5pt}
    \captionof{table}{
	\textbf{Computational resource experiment.} To maintain fairness in the comparison, we consistently set the image size to 224$\times$224. ``Time" means training time.}
	\label{table2}
    \resizebox{0.99\linewidth}{!}{
	\begin{tabular}{lccccc}
    \toprule
    Method  & Flops (GMac) & Params (M) & Time (hour) &21K-val (\%) & Zero-shot (\%)   \\ \midrule
    POMP (our impl.)~\cite{ren2023prompt}  &18.31 & 149.628  & 1.4 & 24.0 & 65.49\\
    RPP (ours)   &18.71 & 149.794  & 1.6 & 24.9 & 65.92\\ 
    \bottomrule
    \end{tabular}
    }
    \vspace{-1em}
\end{wraptable}

%% file: tables/templates.tex
\begin{table}[!h]
\vspace{-1em}
\captionsetup{font={scriptsize}}
\caption{\textbf{Datasets in our experiments.} Text template utilized by senior CLIP teacher for different datasets.}
\label{tab:dataset}
\begin{center}
\begin{adjustbox}{max width=\textwidth}
\begin{tabular}{llrrrr}
\toprule
Dataset  & Text template& Classes & Train Size & Test Size & Metric\\
\midrule
\emph{Prompt Pretraining} \\
ImageNet-21K~\cite{deng2009imagenet}& \textsf{`` a photo of a [class]. "}& 11,221 & 12,358,688 & 561,050 & accuracy\\ 
\midrule
\emph{Datasets of Image Classification} \\
Caltech-101~\cite{FeiFei2004LearningGV} & \textsf{`` a photo of a [class]. "} & 102 & 3,060 & 6,086 & mean per-class accuracy\\
Oxford-Pets~\cite{parkhi2012cats} & \textsf{`` a photo of a [class], a type of pet. "} & 37 & 3,680 & 3,669 & mean per-class accuracy\\
Stanford Cars~\cite{Krause20133DOR} & \textsf{`` a photo of a [class]. "} & 196 & 8,144 & 8,041 & accuracy\\
Oxford Flowers-102~\cite{Nilsback2008AutomatedFC} & \textsf{`` a photo of a [class], a type of flower. "} & 102 & 2,040 & 6,149 & mean per-class accuracy\\
Food-101~\cite{Bossard2014Food101M} & \textsf{`` a photo of [class], a type of food. "} & 101 & 75,750 & 25,250 & accuracy\\
FGVC Aircraft~\cite{Maji2013FineGrainedVC} & \textsf{`` a photo of a [class], a type of aircraft. "} & 100 & 6,667 & 3,333 & mean per-class accuracy\\
SUN-397~\cite{Xiao2010SUNDL}  & \textsf{`` a photo of a [class]. "} & 397 & 15,880 & 19,850 & accuracy \\
Describable Textures (DTD)~\cite{Cimpoi2014DescribingTI}& \textsf{`` [class] texture. "}  & 47 & 3,760 & 1,880 & accuracy\\ 
EuroSAT~\cite{Helber2019EuroSATAN} & \textsf{`` a centered satellite photo of [class]. "}& 10 & 10,000 & 5,000 & accuracy \\
UCF-101~\cite{Soomro2012UCF101AD} & \textsf{`` a photo of a person doing [class]. "} & 101 & 7,639 & 3,783 & accuracy \\
ImageNet~\cite{Deng2009ImageNetAL} & \textsf{`` a photo of a [class]. "}& 1000 & 1,281,167 & 50,000 & accuracy \\
ImageNetV2~\cite{Recht2019DoIC} & \textsf{`` a photo of a [class]. "}& 1,000 & 10,000 & 10,000 & accuracy \\
ImageNet-S~\cite{Wang2019LearningRG} & \textsf{`` a photo of a [class]. "}& 1,000 & 50,889 & 50,889 & accuracy \\
ImageNet-R~\cite{Hendrycks2020TheMF} & \textsf{`` a photo of a [class]. "}& 200 & 30,000 & 30,000 & accuracy \\ 
\bottomrule
\end{tabular}
\end{adjustbox}
\end{center}
\vspace{-1em}
\end{table}

%% file: tables/ablation-study.tex
\begin{table}[]
\captionsetup{font={scriptsize}}
\caption{\textbf{Ablation Study Results.} By default, ablation experiments employ ViT-B/16 as the backbone, with ViT-L/14 serving as the Teacher model. All ablation studies are conducted using the same GPUs. ``Nctx" means the number of prompts that can be learned per layer. ``Pretrained val.” means ImageNet-21K validation Top-1 accuracy. ``Zero-shot" means average accuracy on cross-dataset and cross-domain evaluation for image classification. ``-" indicates that the item is empty. We performed ablation experiments on ImageNet-21K and 14 zero-shot datasets.}
\begin{adjustbox}{max width=\textwidth}
\begin{tabular}{@{}l cc ccc ll @{}}
\toprule
Method &SAPL &PPKD & Nctx &PPKD temperature & PPKD loss-weight  & Pretrained val. (\%)  & Zero-shot (\%)    \\ 
\midrule
\multirow{4}{*}{RPP} 
&- &- & 16 & - & - & 24.0 & 65.49  \\ 
&\checkmark&-& 4 & 1.0 & 0.0 & 24.9 \improve{$\uparrow$} & 64.03 \descend{$\downarrow$} \\
&-&\checkmark& 4 & 1.0 & 1.0 & 24.1 \descend{--} & 65.64 \improve{$\uparrow$} \\
&\checkmark&\checkmark& 4 & 1.0 & 1.0 & 25.2 \improve{$\uparrow$} & 65.70 \improve{$\uparrow$} \\
\midrule
\multirow{14}{*}{Ablation Study}  & \multirow{14}{*}{\checkmark}  & \multirow{14}{*}{\checkmark}
& 8 & \demph{1.0} & \demph{1.0} & 25.4  & 65.59 \\
&&& 16 & \demph{1.0} & \demph{1.0} & 25.5 & 65.16 \\
&&& 32 & \demph{1.0} & \demph{1.0} & 25.6 & 65.17 \\
&&& \demph{4} & 0.1 & \demph{1.0} & 25.0 & 64.76 \\
&&& \demph{4} & 0.2 & \demph{1.0} & 25.1 & 64.66 \\
&&& \demph{4} & 2.0 & \demph{1.0} & 24.9 & 65.66 \\
&&& \demph{4} & 4.0 & \demph{1.0} & 24.7 &  65.27 \\
&&& \demph{4} & 8.0 & \demph{1.0} & 24.6 &  64.96 \\
&&& \demph{4} & \demph{1.0} & 0.1 & 25.1 &  64.54 \\
&&& \demph{4} & \demph{1.0} & 0.5 & 25.2 &  65.28 \\
&&& \demph{4} & \demph{1.0} & 1.0 & 25.2 & 65.70 \\
&&& \demph{4} & \demph{1.0} & 1.5 & 25.0 & 65.41 \\
\multicolumn{3}{c}{\cellcolor{white}} & \cellcolor{tablecolor}\demph{4} & \cellcolor{tablecolor}\demph{1.0} & \cellcolor{tablecolor}2.0 & \cellcolor{tablecolor}24.9 & \cellcolor{tablecolor}65.92 \\ 
&&& \demph{4} & \demph{1.0} & 4.0 & 24.7 & 65.41 \\
\bottomrule
\end{tabular}
\end{adjustbox}
\label{table:ablation-study}
\vspace{-1em}
\end{table}

%% file: tables/base-to-new.tex
\begin{table}[]
    \centering
    \captionsetup{font={scriptsize}}
    \caption{\textbf{The complete experiment from base-to-novel.} Comparison with state-of-the-art methods \textit{w/} or \textit{w/o} RPP on base-to-novel generalization. Our RPP consistently improves baseline model performance on 11 datasets.}
    \vspace{10pt}
    \begin{subtable}[t]{0.32\linewidth}
        \resizebox{1\linewidth}{!}
        {
        \begin{tabular}{cccl}
            \toprule
            Method    & Base (\%)  & Novel (\%) & HM (\%) \\
            \midrule
            CLIP & 69.34 & 74.22 & 71.70 \\
            CoCoOp & 80.47 & 71.69 & 75.83 \\
            MaPLe   & 82.28 & 75.14 & 78.55 \\
            \midrule
            CoOp & 82.69 & 63.22 & 71.66 \\
            + POMP (our impl.) &78.51 	&70.14 	&74.09    \improve{(+2.43)} \\
            \rowcolor{tablecolor} \textbf{+ RPP (ours)} &82.19 	&76.77	&79.39    \improve{(+7.73)} \\
            \midrule
            PromptSRC & 84.26 & 76.10 & 79.97 \\
            + POMP (our impl.) &83.48 	&73.78 	&78.33    \descend{(-1.64)} \\
            \rowcolor{tablecolor} \textbf{+ RPP (ours)} &85.21  &77.38 &81.10  \improve{(+1.13)}   \\
            \bottomrule
        \end{tabular}
        }
        \caption{Average over 11 datasets.}
    \end{subtable}
    \begin{subtable}[t]{0.32\linewidth}
        \resizebox{1\linewidth}{!}
        {
        \begin{tabular}{cccl}
            \toprule
            Method    & Base (\%)  & Novel (\%) & HM (\%) \\
            \midrule
            CLIP & 72.43 & 68.14 & 70.22 \\
            CoCoOp & 75.98 & 70.43 & 73.10 \\
            MaPLe   & 76.66 & 70.54 & 73.47 \\
            \midrule
            CoOp & 76.47 & 67.88 & 71.92 \\
            + POMP (our impl.) &74.43 	&67.07 	&70.55    \descend{(-1.37)} \\
            \rowcolor{tablecolor} \textbf{+ RPP (ours)} &76.77 	&72.50	&74.57    \improve{(+2.65)} \\
            \midrule
            PromptSRC & 77.60 & 70.73 & 74.01 \\
            + POMP (our impl.) &77.53 	&70.47 	&73.83    \descend{(-0.18)} \\
            \rowcolor{tablecolor} \textbf{+ RPP (ours)} &77.93 	&72.50 	&75.12  \improve{(+1.11)} \\
            \bottomrule
        \end{tabular}
        }
        \caption{ImageNet}
    \end{subtable}
    \begin{subtable}[t]{0.32\linewidth}
        \resizebox{1\linewidth}{!}
        {
        \begin{tabular}{cccl}
            \toprule
            Method    & Base (\%)  & Novel (\%) & HM (\%) \\
            \midrule
            CLIP & 96.84 & 94.00 & 95.40 \\
            CoCoOp & 97.96 & 93.81 & 95.84 \\
            MaPLe & 97.74 & 94.36 & 96.02 \\
            \midrule
            CoOp & 98.00 & 89.81 & 93.73 \\
            + POMP (our impl.) &97.93 	&93.27 	&95.54    \improve{(+1.81)} \\ 
            \rowcolor{tablecolor} \textbf{+ RPP (ours)} &98.67 	&95.33	&96.97    \improve{(+3.24)} \\
            \midrule
            PromptSRC & 98.10 & 94.03 & 96.02 \\
            \rowcolor{tablecolor} \textbf{+ POMP (our impl.)} &98.40 	&95.37 	&96.86    \improve{(+0.84)} \\
            + RPP  (ours) &98.67  &95.00 	&96.80   \improve{(+0.78)} \\
            \bottomrule
        \end{tabular}
        }
        \caption{Caltech101}
    \end{subtable}
    
    \vspace{4pt}
    
    \begin{subtable}[t]{0.32\linewidth}
        \resizebox{1\linewidth}{!}
        {
        \begin{tabular}{cccl}
            \toprule
            Method    & Base (\%)  & Novel (\%) & HM (\%) \\
            \midrule
            CLIP & 91.17 & 97.26 & 94.12 \\
            CoCoOp & 95.20 & 97.69 & 96.43 \\
            MaPLe & 95.43 & 97.76 & 96.58 \\
            \midrule
            CoOp & 93.67 & 95.29 & 94.47 \\
            + POMP (our impl.) &95.57 	&97.40 	&96.47    \improve{(+2.00)} \\  
            \rowcolor{tablecolor} \textbf{+ RPP (ours)} &96.17 	&97.77	&96.96    \improve{(+2.49)} \\
            \midrule
            PromptSRC & 95.33 & 97.30 & 96.30 \\
            + POMP (our impl.) &96.43 	&97.43 	&96.93   \improve{(+0.63)} \\
            \rowcolor{tablecolor} \textbf{+ RPP (ours)} &96.37 	&97.93 	&97.14  \improve{(+0.84)}  \\ 
            \bottomrule
        \end{tabular}
        }
        \caption{OxfordPets}
    \end{subtable}
    \begin{subtable}[t]{0.32\linewidth}
        \resizebox{1\linewidth}{!}
        {
        \begin{tabular}{cccl}
            \toprule
            Method    & Base (\%) & Novel (\%) & HM (\%) \\
            \midrule
            CLIP & 63.37 & 74.89 & 68.65 \\
            CoCoOp & 70.49 & 73.59 & 72.01 \\
            MaPLe & 72.94 & 74.00 & 73.47 \\
            \midrule
            CoOp & 78.12 & 60.40 & 68.13 \\
            + POMP (our impl.) &71.27 	&71.90 	&71.58   \improve{(+3.45)} \\   
            \rowcolor{tablecolor} \textbf{+ RPP (ours)} &75.13 	&75.77	&75.45   \improve{(+7.32)} \\
            \midrule
            PromptSRC & 78.27 & 74.97 & 76.58 \\
            + POMP (our impl.) &77.07 	&74.30 	&75.66   \descend{(-0.92)} \\ 
            \rowcolor{tablecolor} \textbf{+ RPP (ours)} &80.87 	&75.83 	&78.27  \improve{(+1.69)} \\
            \bottomrule
        \end{tabular}
        }
        \caption{StanfordCars}
    \end{subtable}
    \begin{subtable}[t]{0.32\linewidth}
        \resizebox{1\linewidth}{!}
        {
        \begin{tabular}{cccl}
            \toprule
            Method    & Base (\%)  & Novel (\%) & HM (\%)\\
            \midrule
            CLIP & 72.08 & 77.80 & 74.83 \\
            CoCoOp & 94.87 & 71.75 & 81.71 \\
            MaPLe & 95.92 & 72.46 & 82.56 \\
            \midrule
            CoOp & 97.60 & 59.67 & 74.06 \\
            + POMP (our impl.) &93.43 	&72.60 	&81.71   \improve{(+7.65)} \\ 
            \rowcolor{tablecolor} \textbf{+ RPP (ours)} &96.67 	&77.60	&86.09   \improve{(+12.03)} \\
            \midrule
            PromptSRC & 98.07 & 76.50 & 85.95 \\
            + POMP (our impl.) &97.37 	&73.67 	&83.87   \descend{(-2.08)} \\
            \rowcolor{tablecolor} \textbf{+ RPP (ours)} &98.00	 &77.73 	&86.70  \improve{(+0.75)} \\ 
            \bottomrule
        \end{tabular}
        }
        \caption{Flowers102}
    \end{subtable}

    \vspace{4pt}
    
    \begin{subtable}[t]{0.32\linewidth}
        \resizebox{1\linewidth}{!}
        {
        \begin{tabular}{cccl}
            \toprule
            Method    & Base (\%)  & Novel (\%) & HM (\%)\\
            \midrule
            CLIP & 90.10 & 91.22 & 90.66 \\
            CoCoOp & 90.70 & 91.29 & 90.99 \\
            MaPLe & 90.71 & 92.05 & 91.38 \\
            \midrule
            CoOp & 88.33 & 82.26 & 85.19 \\
            + POMP (our impl.) &89.73 	&90.87 	&90.30   \improve{(+5.11)} \\ 
            \rowcolor{tablecolor} \textbf{+ RPP (ours)} &90.07 	&91.53	&90.79   \improve{(+5.60)} \\
            \midrule
            PromptSRC & 90.67 & 91.53 & 91.10 \\
            + POMP (our impl.) &90.50 	&91.67 	&91.08   \descend{(-0.02)} \\ 
            \rowcolor{tablecolor} \textbf{+ RPP (ours)} &90.70 	&92.07 	&91.38  \improve{(+0.28)}  \\
            \bottomrule
        \end{tabular}
        }
        \caption{Food101}
    \end{subtable}
    \begin{subtable}[t]{0.32\linewidth}
        \resizebox{1\linewidth}{!}
        {
        \begin{tabular}{cccl}
            \toprule
            Method    & Base (\%) & Novel (\%) & HM (\%) \\
            \midrule
            CLIP & 27.19 & 36.29 & 31.09 \\
            CoCoOp & 33.41 & 23.71 & 27.74 \\
            MaPLe & 37.44 & 35.61 & 36.50 \\
            \midrule
            CoOp & 40.44 & 22.30 & 28.75 \\
            + POMP (our impl.) &21.43 	&23.77 	&22.54   \descend{(-6.21)} \\ 
            \rowcolor{tablecolor} \textbf{+ RPP (ours)} &37.07 	&36.10	&36.58   \improve{(+7.83)} \\
            \midrule
            PromptSRC & 42.73 & 37.87 & 40.15 \\
            + POMP (our impl.) &33.90 	&28.63 	&31.04   \descend{(-9.11)} \\ 
            \rowcolor{tablecolor} \textbf{+ RPP (ours)} &43.73 	&38.20	&40.78   \improve{(+0.63)} \\ 
            \bottomrule
        \end{tabular}
        }
        \caption{FGVCAircraft}
    \end{subtable}
    \begin{subtable}[t]{0.32\linewidth}
        \resizebox{1\linewidth}{!}
        {
        \begin{tabular}{cccl}
            \toprule
            Method    & Base (\%) & Novel (\%) & HM (\%)\\
            \midrule
            CLIP & 69.36 & 75.35 & 72.23 \\
            CoCoOp & 79.74 & 76.86 & 78.27 \\
            MaPLe & 80.82 & 78.70 & 79.75 \\
            \midrule
            CoOp & 80.60 & 65.89 & 72.51 \\
            + POMP (our impl.) &79.70 	&74.80 	&77.17  \improve{(+4.66)} \\ 
            \rowcolor{tablecolor} \textbf{+ RPP (ours)} &81.63 	&79.20	&80.40   \improve{(+7.89)} \\
            \midrule
            PromptSRC & 82.67 & 78.47 & 80.52 \\ 
            + POMP (our impl.) &82.37 	&78.53 	&80.40  \descend{(-0.12)} \\ 
            \rowcolor{tablecolor} \textbf{+ RPP (ours)} &82.87 	&79.20 	&80.99  \improve{(+0.47)}  \\
            \bottomrule
        \end{tabular}
        }
        \caption{SUN397}
    \end{subtable}

    \vspace{4pt}

    \begin{subtable}[t]{0.32\linewidth}
        \resizebox{1\linewidth}{!}
        {
        \begin{tabular}{cccl}
            \toprule
            Method    & Base (\%) & Novel (\%)& HM (\%)\\
            \midrule
            CLIP & 53.24 & 59.90 & 56.37 \\
            CoCoOp & 77.01 & 56.00 & 64.85 \\
            MaPLe & 80.36 & 59.18 & 68.16 \\
            \midrule
            CoOp & 79.44 & 41.18 & 54.24 \\ 
            + POMP (our impl.) &76.97 	&44.90 	&56.71  \improve{(+2.47)} \\  
            \rowcolor{tablecolor} \textbf{+ RPP (ours)} &81.23 	&59.37	&68.60   \improve{(+14.36)} \\
            \midrule
            PromptSRC & 83.37 & 62.97 & 71.75 \\
            + POMP (our impl.) &83.57 	&50.63 	&63.06  \descend{(-8.69)} \\ 
            \rowcolor{tablecolor} \textbf{+ RPP (ours)} &84.47     &64.73  &72.65   \improve{(+0.90)} \\ 
            \bottomrule
        \end{tabular}
        }
        \caption{DTD}
    \end{subtable}
    \begin{subtable}[t]{0.32\linewidth}
        \resizebox{1\linewidth}{!}
        {
        \begin{tabular}{cccl}
            \toprule
            Method    & Base (\%) & Novel (\%)& HM (\%)\\
            \midrule
            CLIP & 56.48 & 64.05 & 60.03 \\
            CoCoOp & 87.49 & 60.04 & 71.21 \\
            MaPLe & 94.07 & 73.23 & 82.35 \\
            \midrule
            CoOp & 92.19 & 54.74 & 68.69 \\
            + POMP (our impl.) &80.83 	&63.10 	&70.87  \improve{(+2.18)} \\
            \rowcolor{tablecolor} \textbf{+ RPP (ours)} &86.20 	&81.37	&83.71   \improve{(+15.02)} \\
            \midrule
            PromptSRC & 92.90 & 73.90 & 82.32 \\
            + POMP (our impl.) &94.27 	&72.97 	&82.26  \descend{(-0.06)} \\ 
            \rowcolor{tablecolor} \textbf{+ RPP (ours)} &96.27 	&79.37 	&87.01   \improve{(+4.69)} \\
            \bottomrule
        \end{tabular}
        }
        \caption{EuroSAT}
    \end{subtable}
    \begin{subtable}[t]{0.32\linewidth}
        \resizebox{1\linewidth}{!}
        {
        \begin{tabular}{cccl}
            \toprule
            Method    &  Base (\%) & Novel (\%)& HM (\%)\\
            \midrule
            CLIP & 70.53 & 77.50 & 73.85 \\
            CoCoOp & 82.33 & 73.45 & 77.64 \\
            MaPLe & 83.00 & 78.66 & 80.77 \\
            \midrule
            CoOp & 84.69 & 56.05 & 67.46 \\
            + POMP (our impl.) &82.27 	&71.87 	&76.72  \improve{(+9.26)} \\
            \rowcolor{tablecolor} \textbf{+ RPP (ours)} &84.50 	&77.97	&81.10   \improve{(+13.64)} \\ 
            \midrule
            PromptSRC & 87.10 & 78.80 & 82.74 \\
            + POMP (our impl.) &86.90 	&77.90 	&82.15 \descend{(-0.59)} \\ 
            \rowcolor{tablecolor} \textbf{+ RPP (ours)} &87.43 	&79.57 	&83.31   \improve{(+0.57)} \\ 
            \bottomrule
        \end{tabular}
        }
        \caption{UCF101}
    \end{subtable}
    \label{table:base-to-novel}
    \vspace{-2em}
\end{table}

%% file: tables/few-shot-allshot.tex
\begin{table}[]
    \captionsetup{font={scriptsize}}
    \caption{\textbf{The performance of RPP (base on PromptSRC) and compared methods in few-shot setting.}}
    \begin{adjustbox}{max width=0.98\textwidth}
    \setlength{\tabcolsep}{15pt}
    \scalebox{0.99}{
    \begin{tabular}{llccccc}
    \toprule
    Dataset & 
    Method & 
    1 shot (\%)&
    2 shots (\%)& 
    4 shots (\%)& 
    8 shots (\%)&
    16 shots (\%)\\  
    \midrule
    \multirow{5}{*}{ImageNet}      & Linear probe CLIP      &32.13	&44.88&	54.85	&62.23	&67.31\\ 
                                   & CoOp                   &66.33	&67.07	&68.73	&70.63&	71.87\\ 
                                   & CoCoOp                  & 69.43	&69.78	&70.39&	70.63&	70.83\\
                                   & MaPLe                  & 62.67	&65.10	&67.70&	70.30&	72.33\\
                                   & PromptSRC             & 68.13	&69.77	&71.07	&72.33	&73.17\\
                                   \rowcolor{tablecolor}
                                   \multicolumn{1}{c}{\cellcolor{white}}& RPP \textbf{(ours) }     &71.83   &72.80  &73.40    &73.87   & 73.87    \\
    \midrule
    \multirow{5}{*}{Caltech101}    & Linear probe CLIP      & 79.88	&89.01	&92.05	&93.41	&95.43\\
                                   & CoOp                    & 92.60	&93.07	&94.40	&94.37	&95.57\\
                                   & CoCoOp                  & 93.83	&94.82	&94.98	&95.04	&95.16\\
                                   & MaPLe                  & 92.57	&93.97	&94.43&	95.20&	96.00\\
                                   &  PromptSRC              &93.67	&94.53	&95.27	&95.67	&96.07\\
                                   \rowcolor{tablecolor}
                                   \multicolumn{1}{c}{\cellcolor{white}}& RPP \textbf{(ours) }    &95.43 	&95.77 	&96.20 	&96.37 	&96.77   \\
    \midrule
    \multirow{5}{*}{DTD}           & Linear probe CLIP       & 34.59	&40.76	&55.71	&63.46	&69.96\\
                                   & CoOp                   &50.23&	53.60	&58.70	&64.77	&69.87\\
                                   & CoCoOp                      & 48.54	&52.17	&55.04	&58.89	&63.04\\
                                   & MaPLe                  & 52.13	&55.50	&61.00&	66.50&	71.33\\
                                   &  PromptSRC                  & 56.23&59.97	&65.53	&69.87	&72.73\\
                                   \rowcolor{tablecolor}
                                    \multicolumn{1}{c}{\cellcolor{white}}& RPP \textbf{(ours) }    &63.73  &70.43 &72.87 &75.47 	&73.60     \\
    \midrule
    \multirow{5}{*}{EuroSAT}       & Linear probe CLIP       &49.23	&61.98	&77.09&	84.43	&87.21\\
                                   & CoOp                   &54.93	&65.17	&70.80	&78.07	&84.93\\
                                   & CoCoOp                  & 55.33	&46.74	&65.56&	68.21	&73.32\\
                                    & MaPLe                  & 71.80	&78.30	&84.50&	87.73&	92.33\\
                                   & PromptSRC              &73.13 &79.37	&86.30	&88.80	&92.43\\
                                   \rowcolor{tablecolor}
                                    \multicolumn{1}{c}{\cellcolor{white}}& RPP \textbf{(ours) }   &80.4   &82.03 	&90.67 	&93.73 	&93.37     \\
    \midrule
    \multirow{5}{*}{StanfordCars}  & Linear probe CLIP     & 35.66	&50.28	&63.38	&73.67	&80.44\\
                               & CoOp                        &67.43	&70.50	&74.47	&79.30	&83.07\\
                                       & CoCoOp              & 67.22	&68.37	&69.39	&70.44&	71.57\\
                                     & MaPLe                  & 66.60	&71.60	&75.30&	79.47&	83.57\\
                                  &  PromptSRC               &69.40	&73.40	&77.13	&80.97	&83.83\\
                                  \rowcolor{tablecolor}
                                   \multicolumn{1}{c}{\cellcolor{white}}& RPP \textbf{(ours) }    &74.10 	&77.60 	&80.73 	&83.47 	&84.87     \\
    \midrule
    \multirow{5}{*}{Flowers102}    & Linear probe CLIP      & 69.74	&85.07	&92.02	&96.10	&97.37\\
                                   & CoOp                   & 77.53	&87.33	&92.17	&94.97	&97.07\\
                                   & CoCoOp                    &72.08	&75.79	&78.40	&84.30	&87.84\\
                                   & MaPLe                  & 83.30	&88.93	&92.67&	95.80&	97.00\\
                                   & PromptSRC            & 85.93	&91.17	&93.87	&96.27	&97.60\\
                                   \rowcolor{tablecolor}
                                    \multicolumn{1}{c}{\cellcolor{white}}& RPP \textbf{(ours) }  &86.67 &92.27  &94.77  &96.60  &97.80     \\
    \midrule
    \multirow{5}{*}{FGVCAircraft}  & Linear probe CLIP       & 19.61	&26.41	&32.33	&39.35&	45.36\\
                                   & CoOp                     & 21.37&	26.20	&30.83	&39.00	&43.40\\
                                   & CoCoOp                   & 12.68	&15.06	&24.79	&26.61	&31.21\\
                                   & MaPLe                  & 26.73	&30.90	& 34.87&	42.00&	48.40\\
                                   & PromptSRC           & 27.67	&31.70	&37.47&	43.27	&50.83\\
                                   \rowcolor{tablecolor}
                                     \multicolumn{1}{c}{\cellcolor{white}}& RPP \textbf{(ours) }   &36.57 &39.47 	&43.60 	&49.10 	&51.77  \\
    \midrule
    \multirow{5}{*}{SUN397}        & Linear probe CLIP          & 41.58	&53.70	&63.00	&69.08	&73.28\\
                                   & CoOp                     & 66.77	&66.53	&69.97	&71.53	&74.67\\
                                   & CoCoOp                   & 68.33	&69.03&	70.21	&70.84	&72.15\\
                                   & MaPLe                  & 64.77	&67.10	&70.67&	73.23&	75.53\\
                                  & PromptSRC                &69.67	&71.60	&74.00	&75.73	&77.23\\
                                  \rowcolor{tablecolor}
                                   \multicolumn{1}{c}{\cellcolor{white}}& RPP \textbf{(ours) }   & 73.23 	&74.57 	&75.93 	&77.03  &77.40      \\
    \midrule
    \multirow{5}{*}{OxfordPets}    & Linear probe CLIP        & 44.06	&58.37	&71.17	&78.36	&85.34\\
                                   & CoOp                      & 90.37	&89.80	&92.57	&91.27	&91.87\\
                                   & CoCoOp                       & 91.27	&92.64	&92.81	&93.45	&93.34\\
                                   & MaPLe                  & 89.10	&90.87	&91.90&	92.57& 92.83\\
                                   & PromptSRC                   & 92.00	&92.50	&93.43	&93.50	&93.67\\
                                   \rowcolor{tablecolor}
                                    \multicolumn{1}{c}{\cellcolor{white}}& RPP \textbf{(ours) }    & 92.73 	&93.23 	&93.90 &93.93 &94.47      \\
    \midrule
    \multirow{5}{*}{UCF101}        & Linear probe CLIP        & 53.66	&65.78	&73.28	&79.34&	82.11\\
                                   & CoOp                     & 71.23	&73.43	&77.10	&80.20	&82.23\\
                                   & CoCoOp                   & 70.30	&73.51	&74.82	&77.14&	78.14\\
                                   & MaPLe                  & 71.83	&74.60	& 78.47& 81.37&	85.03\\
                                   & PromptSRC                   & 74.80	&78.50	&81.57	&84.30	&86.47\\
                                   \rowcolor{tablecolor}
                                    \multicolumn{1}{c}{\cellcolor{white}}& RPP \textbf{(ours) }     &80.67 	&83.23 	&85.23 	&86.63 	&86.50     \\
    \midrule
    \multirow{5}{*}{Food101}       & Linear probe CLIP       & 43.96	&61.51	&73.19	&79.79	&82.90\\
                                   & CoOp                    & 84.33	&84.40	&84.47	&82.67	&84.20\\
                                   & CoCoOp                     & 85.65	&86.22	&86.88	&86.97	&87.25\\
                                   & MaPLe                  & 80.50 &81.47	&81.77&	83.60&	85.33\\
                                   & PromptSRC                & 84.87&	85.70	&86.17	&86.90	&87.5\\
                                   \rowcolor{tablecolor}
                                    \multicolumn{1}{c}{\cellcolor{white}}& RPP \textbf{(ours) }   &86.23 &86.57 	&86.70 	&87.07 	&87.53      \\
    \midrule
    \multirow{5}{*}{Average}       & Linear probe CLIP       & 45.83	&57.98	&68.01	&74.47&	78.79\\
                                   & CoOp                    & 67.56	&70.65	&74.02	&76.98	&79.89\\
                                   & CoCoOp                & 66.79&	67.65	&71.21&	72.96	&74.90\\
                                   & MaPLe                  & 69.27	&72.58	&75.37&	78.89&	81.79\\
                                   & PromptSRC               & 72.32	&75.29	&78.35	&80.69	&82.87\\
                                   \rowcolor{tablecolor} 
                                   \multicolumn{1}{c}{\cellcolor{white}}& RPP \textbf{(ours) } &  76.51 & 78.91  & 81.27 & 83.02 & 83.45   \\
    
    \bottomrule
    \end{tabular}
    }
    \end{adjustbox}
    \label{tab_appendix:few_shot_experiments}
    \vspace{-5mm}
\end{table}